\documentclass{article} 

\usepackage[hidelinks]{hyperref}
\usepackage{subfigure}
\usepackage{fancyvrb}
\usepackage{amsmath, amsfonts, amsthm, amssymb}
\usepackage{stmaryrd}
\usepackage{verbatim}
\usepackage{graphicx}
\usepackage{setspace}
\usepackage[ruled, vlined]{algorithm2e}
\graphicspath{{.}}

\usepackage{listings}
\usepackage{xcolor}
\definecolor{codegreen}{rgb}{0,0.6,0}
\definecolor{codegray}{rgb}{0.5,0.5,0.5}
\definecolor{codepurple}{rgb}{0.58,0,0.82}
\definecolor{backcolour}{rgb}{0.95,0.95,0.92}

\lstdefinestyle{mystyle}{
    backgroundcolor=\color{backcolour},   
    commentstyle=\color{codegreen},
    keywordstyle=\color{magenta},
    numberstyle=\tiny\color{codegray},
    stringstyle=\color{codepurple},
    basicstyle=\ttfamily\footnotesize,
    breakatwhitespace=false,         
    breaklines=true,                 
    captionpos=b,                    
    keepspaces=true,                 
    numbers=left,                    
    numbersep=5pt,                  
    showspaces=false,                
    showstringspaces=false,
    showtabs=false,                  
    tabsize=2
}
\newtheorem{theorem}{Theorem}[section]

\theoremstyle{definition}
\newtheorem{definition}[theorem]{Definition}
\newtheorem{remark}[theorem]{Remark}

\numberwithin{equation}{section}

\title{Optimizing Transformer Neural Network for Real-Time Outlier Detection on FPGAs}
\author{Ilia Sobakinskikh and Paul Alexander Bilokon}

\usepackage{tikz}
\usetikzlibrary{positioning}

\begin{document}

\maketitle

\begin{abstract}
In this work, we explore how the inference time of a Transformer Neural Network can be efficiently optimized with applications to real-time anomaly detection in financial time series.
The financial time series are price series such as asset prices.
Unfortunately, the data is often with errors or outliers that make the downstream data processing tasks useless, unstable or even harmful \cite{Falkenberry_2008} \cite{Vallis_Hochenbaum_Twitter}. Moreover, the amount of financial time-series data has been significantly increasing \cite{AnomalyDataBig}.
Hence, there is a need for better data-cleaning methods in terms of accuracy and in terms of processing speed.

Transformers as a neural network architecture have achieved superior performances in many tasks such as Natural Language Processing and Computer Vision \cite{TransformersNLP}.
Time series modelling and especially anomaly detection tasks can benefit from the features of transformers architecture in multiple ways, including the capacity to capture long-range dependencies and interactions \cite{2202.07125}.

Increasingly powerful hardware, such as field-programmable gate arrays (FPGAs), have seen increasing usage in recent years due to their reconfigurability and high performance \cite{10.1007/978-3-319-56258-2_14}.
They can be efficiently utilized to speed up the computations of the Transformer architecture.

We explore different Transformer architectures for time series modelling and how they can be efficiently implemented on an FPGA board (PYNQ-Z2).
In particular, we examine the application of Transformers to detect anomalies
in time series and we show how they can be efficiently implemented on an FPGA board
to minimize latency.

The code is available at \url{https://github.com/thxi/icl_thesis}
\end{abstract}

\section{Introduction}

The rapid growth of financial time series data has given rise to significant
challenges in maintaining data quality and processing speed. In this study, we address
these challenges by proposing an implementation of Transformer-based Neural Networks
which achieve good performance in anomaly detection tasks while having
a low inference time.

Financial time series data often suffer from data inaccuracies, errors, and
outliers, rendering downstream data processing tasks ineffective or even detrimental to
decision-making processes. The urgency of this issue is underscored by the ever-increasing
volume of financial time-series data available. Consequently, there is a pressing
demand for more accurate and faster data-cleaning methods.

Transformers, as a powerful neural network architecture, have consistently
demonstrated exceptional performance across various domains, including Natural Language
Processing (NLP) and Computer Vision. Leveraging the capabilities of Transformer
architectures for time series modeling, especially in the realm of anomaly detection, offers
numerous advantages, including the ability to capture long-range dependencies and
intricate interactions within the data.

The evolution of hardware, for example, the growing utilization of
field-programmable gate arrays (FPGAs), has brought about increased computational capabilities
due to their reconfigurability, high performance and deterministic latency. These FPGAs can be
effectively used to accelerate computations associated with Transformer
architectures.

\subsection{Outline}

In Section~\ref{sec:problem_statement}, we will describe the problem statement of anomaly detection task.
In Section~\ref{sec:transformers}, we will describe the main concepts and ideas used in the Transformer architecture.
In Section~\ref{sec:fpga_design}, we will describe the main concepts of programming an FPGA and the specific optimizations that can be applied to speed up the computations.
In Section~\ref{sec:experiments}, we will describe the experiments that we conducted to evaluate the performance of the proposed architectures.
Section~\ref{sec:conclusion} will conclude this work and will provide some ideas for future work.

\subsection{Contributions}

The main contributions of this paper are as follows:
\begin{enumerate}
    \item Implementation of vanilla Transformer architecture on an FPGA board.
    \item Implementation of a Linear Attention Transformer on an FPGA board.
    \item Analysis of the latency/resource utilization tradeoff for the proposed architectures.
    \item Comparative analysis of the performance, related to accuracy, of the proposed architectures on the anomaly detection task.
\end{enumerate}

\section{Problem statement} \label{sec:problem_statement}

In this section, we will describe the problem statement of anomaly detection task.

\subsection{Definitions and examples}

\begin{definition}
    We consider a \textbf{time-series} $\mathcal{T}$ which is simply a timestamped sequence of observations $x_i \in R^n$,
    potentially with some metadata $y_i$.
\end{definition}

Most of the times we will consider univariate case, i.e. $n=1$.
An example of this is a price time-series of a single stock.
However, the multivariate case is also important and we will consider it in the experiments.
For example, one can consider a time-series of prices of multiple stocks to get a multivariate time series or
one can also create features (for example, by calculating the \textbf{differenced time series})
on the original time-series to get a multivariate time-series.

\begin{definition}
    The \textbf{differenced time series} is a time-series $\mathcal{T}$ where each observation $x_i$ is simply the difference
    between the current and the previous observation.
    That is, $x_i=x_{i-1}-x_i$.
\end{definition}

\begin{figure}[h!]
    \centering
    \includegraphics[width=\textwidth]{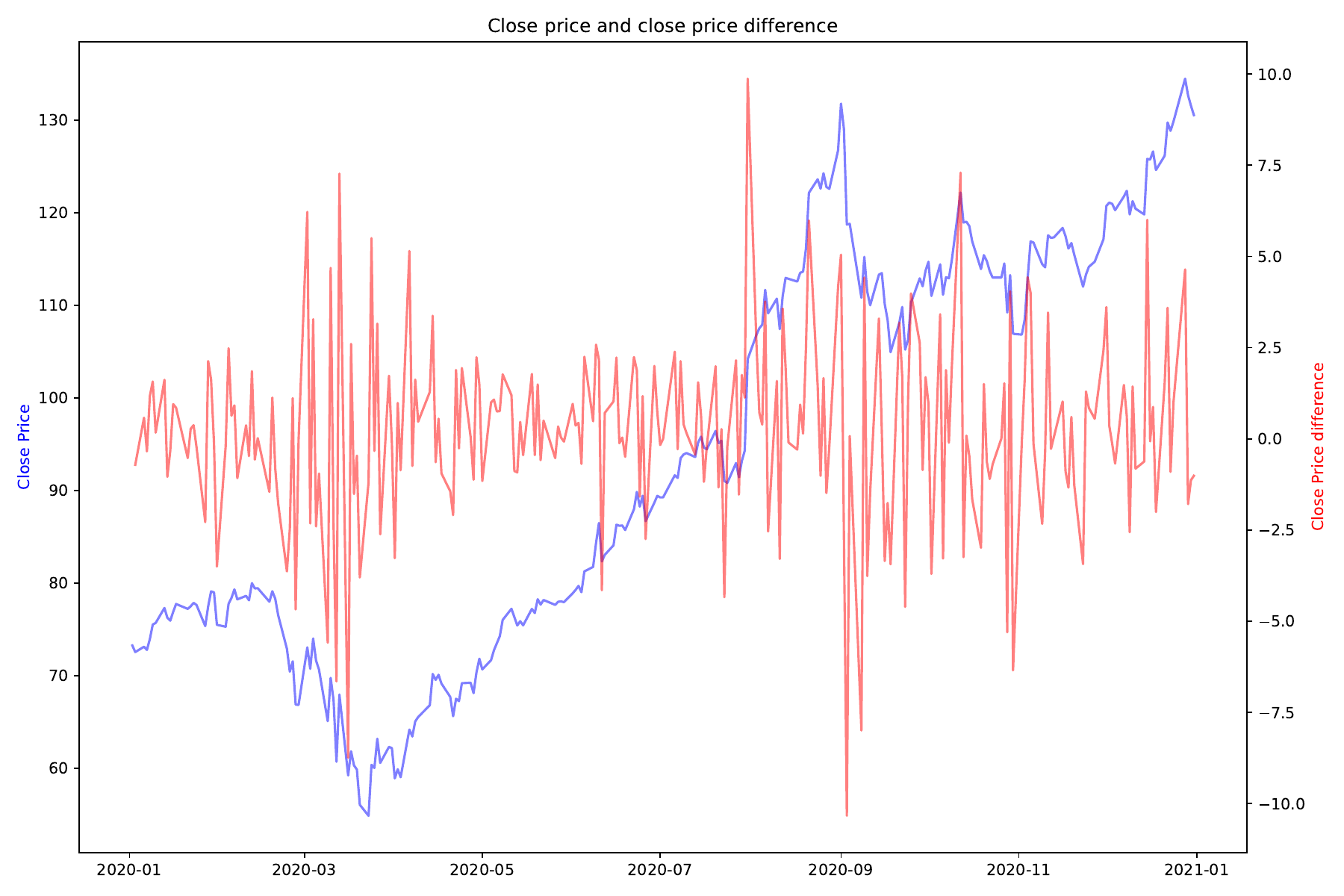}
    \caption{Sample mutli-variate time-series of Apple stock price. The blue line are the close prices
        and the red line is the differenced close prices.}
    \label{fig:apple_time_series}
\end{figure}

\begin{definition}
    The \textbf{Anomaly Detection} task:
    for any time-series $\hat{\mathcal{T}}$ of length $n$, we need to predict $\mathcal{Y} = \{y_1, . . . , y_n \}, y_i \in \{0, 1\}$,
    whether the datapoint at the $i$-th timestamp anomalous (where by convention we will use $1$ as anomaly and $0$ as not an anomaly).
\end{definition}
In this work, we will restrict ourselves to the \textbf{supervised case}
where the labels $y_i$ are known for the \emph{seen} (or training) part of the dataset.

\begin{remark}
    One can also consider an unsupervised task.
    However, one issue with the unsupervised task is that it is hard to evaluate
    the performance (i.e., accuracy) of the model's predictions \cite{1905.05667}.
\end{remark}

\subsection{Different Anomalies Types}

Although there is no universally accepted definition of an anomaly,
there exists a common classification of \textit{synthetic} anomalies into different types. For a broad comprehensive review,
reader is referred to \cite{Steinbuss2020GeneratingAO}, \cite{Schmidl2022AnomalyDI}.

In this work, we will only consider \textbf{Point anomalies} types since
the main focus of this work is not to show the superior performance of the Transformer architecture
on different types of anomalies
but to show that it can be efficiently implemented on an FPGA board.

\begin{definition}
    \textbf{Point anomalies} are individual data points that are considered anomalous with respect to the rest of the data.
    For example, a point anomaly can be a sudden spike in the price of a stock. See Figure \ref{fig:apple_time_series}.

    Mathematically, we could define a point anomaly as follows:
    \begin{equation}
        x_i = \begin{cases}
            x^{\text{original}}_i + S & y_i=1            \\
            x^{\text{original}}       & \text{otherwise}
        \end{cases}
    \end{equation}
    where $x^{\text{original}}$ is the original time-series (as if there was no outliers)
    and $S$ is the spike.

\end{definition}

\begin{figure}[h!]
    \centering
    \includegraphics[width=\textwidth]{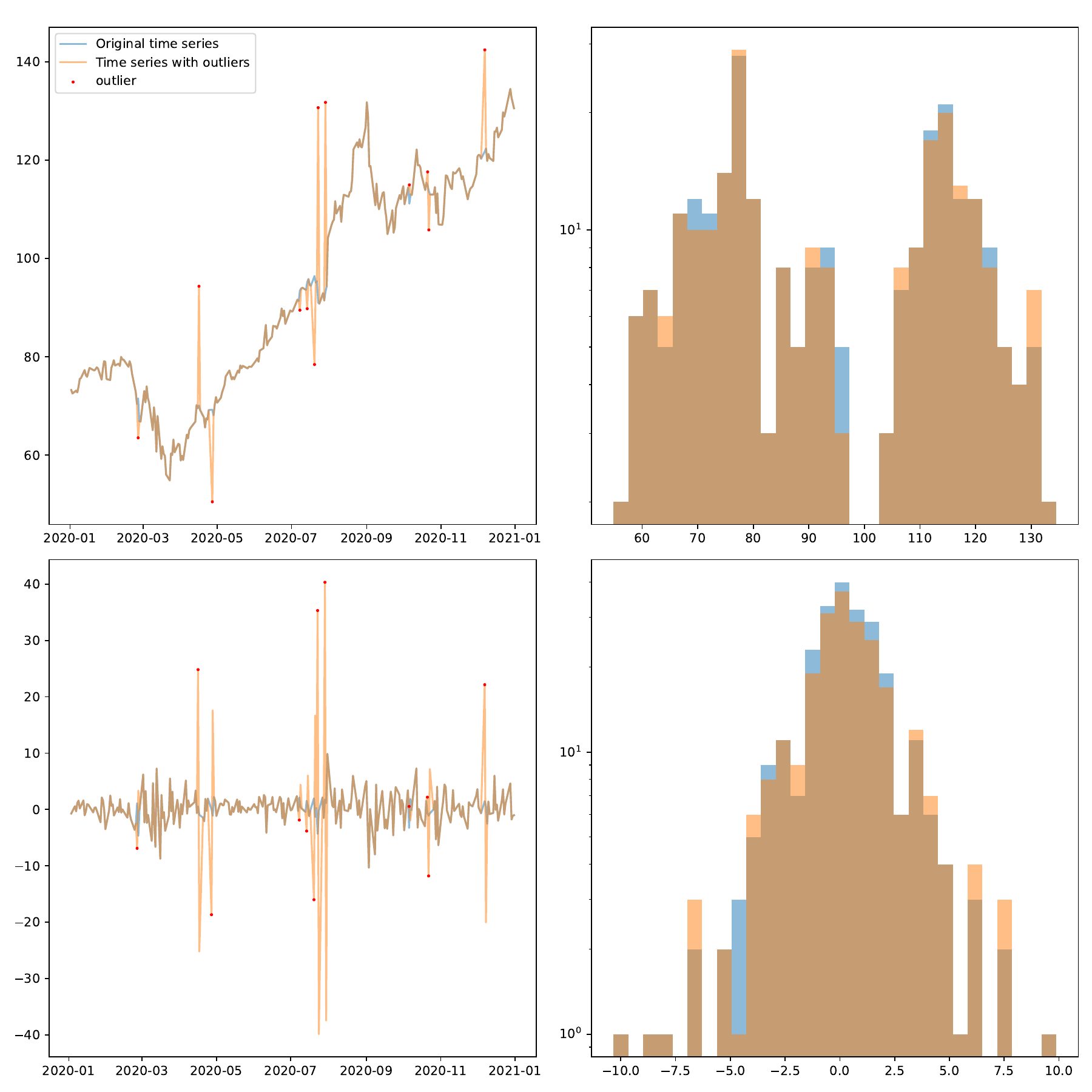}
    \caption{Apple stock modified time series with injected \textbf{Point anomalies}.
        Top panel: price time series and its distribution.
        Bottom panel: differenced price time series and its distribution.}
    \label{fig:apple_time_series_outliers}
\end{figure}

\section{Transformers} \label{sec:transformers}

In this section, we will describe the main concepts and ideas used in the Transformer architecture.
and how to use them for anomaly detection time series task.
We will describe the main building blocks of the Transformer architecture
and will give a special treatment to the attention mechanism firstly introduced in \cite{1409.0473}.

\subsection{General architecture} \label{sec:general_architecture}

In \cite{1706.03762}, authors introduced the \textit{vanilla} Transformer architecture which a neural network architecture
which is the architecture that is dominantly used in Natural Language Processing tasks.
The architecture's main feature was reliance on the attention mechanism and the complete elimination
of recurrent and convolutional layers.

\begin{figure}[h]
    \centering
    \includegraphics[scale=0.5]{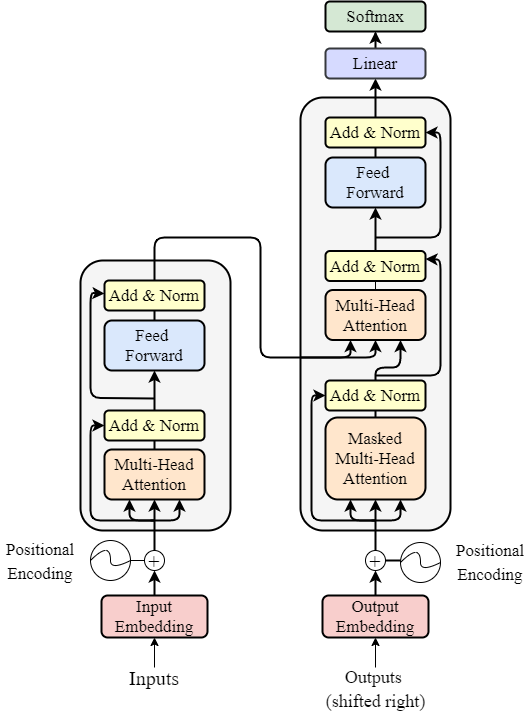}
    \caption{Model architecture of the Transformer \cite{1706.03762}}
    \label{fig:attention_is_all_you_need}
\end{figure}

Figure \ref{fig:attention_is_all_you_need} presents the main architecture of the transformer.
The architecture consists of an \textbf{encoder} and a \textbf{decoder} blocks.
For the purpose of this work we will only consider the \textbf{encoder} part of the architecture
(and in that regard the idea is similar to BERT model for language processing \cite{Devlin2019BERTPO}
which only uses a transformer encoder and this work which uses BERT for anomaly detection \cite{Jeong2023AnomalyBERTST}).
The \textbf{encoder} is preceded by a \textbf{positional encoding} layer which is used to \emph{inject} the positional information
to the input vectors $x_i$ because the attention mechanism is permutation invariant,
this will be explained in Section~\ref{sec:attention_mechanism}.

\subsubsection{Positional encoding}

Since the attention mechanism is permutation invariant
(i.e., it does not take into account the order of the input, refer to Section~\ref{sec:attention_mechanism}),
we need to inject the positional information into the input vectors $x_i$.
In short, a positional encoding is usually a vector of the same dimension as the input vectors $x_i$
which can either be fixed (for example, we have the same vector for different positions) or learned.

For the comprehensive overview of different positional encoding techniques, reader
is referred to \cite{Dufter2021PositionII} and \cite{Weng_2023}.

\begin{figure}[h]
    \centering
    \includegraphics[width=\textwidth]{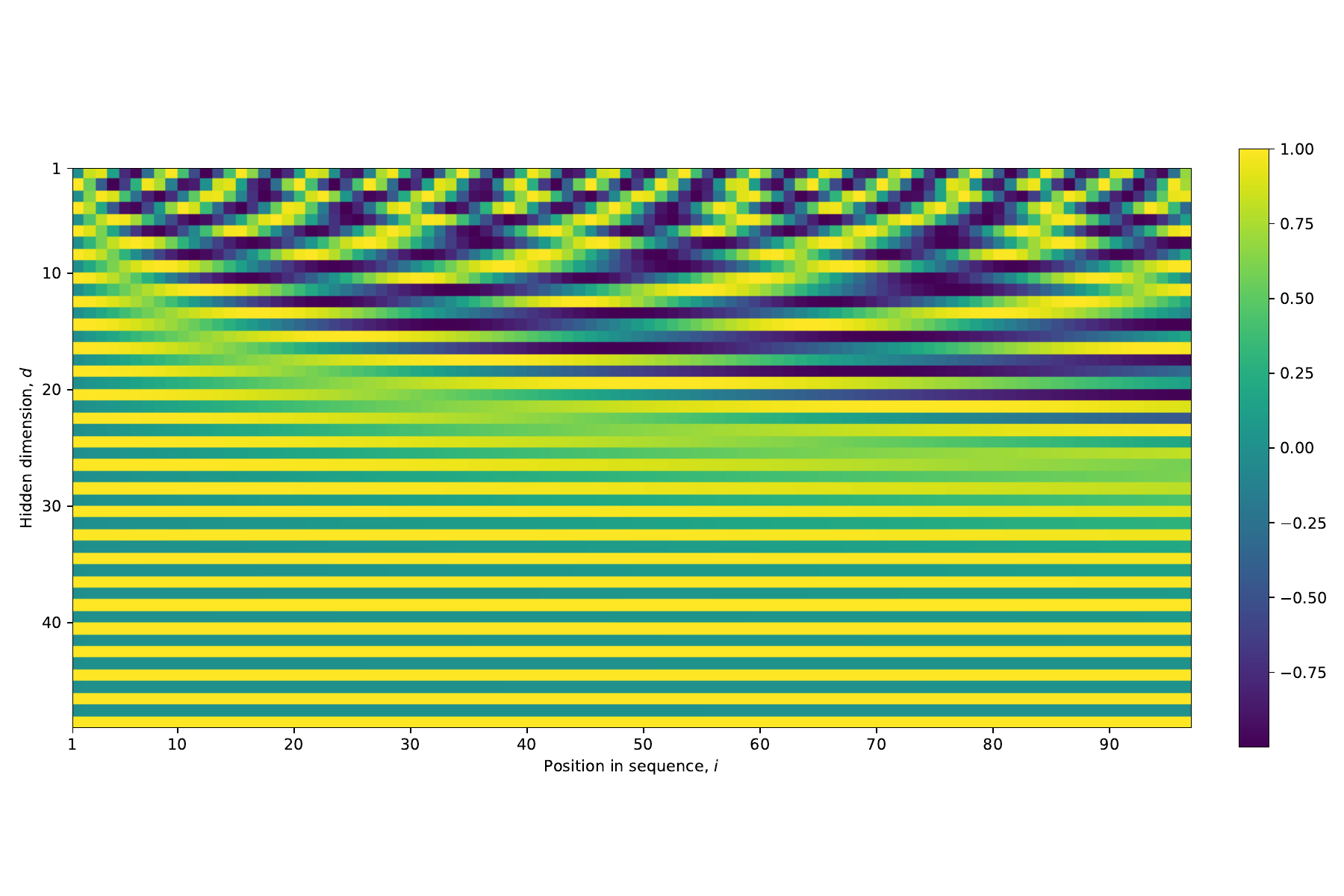}
    \caption{Positional encoding example embeddings for a sequence of max length of 100 and embedding dimension of 48.
        The embeddings are generated using the sinusoidal positional encoding technique.
        The columns represent the embeddings that we would add to the input vector $x_i$.
    }
    \label{fig:pos_encoding}
\end{figure}

The vanilla Transformer architecture uses a sinusoidal positional encoding technique
which is a fixed positional encoding scheme. Suppose we have a time-series of length $T$.
Then the positional encoding for the $i$-th input vector $x_i$ of dimension $d$ is defined as follows:

\begin{equation}
    \begin{array}{rll}
        PE_{i, 2j}   & = \sin(\frac{i}{10000^{2j/d}}) \\
        PE_{i, 2j+1} & = \cos(\frac{i}{10000^{2j/d}})
    \end{array}
\end{equation}
Refer to Figure \ref{fig:pos_encoding} for an example of the embedding vectors that
are obtained using this method.

This combination of sine and cosine functions allows the positional encoding to
generate the embedding of a position in a unique way. The embedding is then
added to the input vector $x_i$, however, it can also be simply stacked with the input vector $x_i$
to get a vector of dimension $2d$. In this work, we will use the former approach.

\subsubsection{Encoder Block} \label{sec:encoder_block}

The encoder consists of $N$ identical layers. Each layer has two sub-layers which are a
\textbf{multi-head self-attention} layer (described in detail in Section~\ref{sec:dot_product_attention})
and a \textbf{feed-forward (FF)} layer.
The \textbf{feed-forward (FF)} layer $\text{FFN}(\cdot)$ is a simple neural network
which comprises of 2 fully-connected ($\text{FC}$) linear layers and an activation function in between them.
Specifically, authors of \cite{1706.03762} used an $\text{FC}$ layer with
the ReLU activation function $$\text{ReLU}(x)=\max(0, x)$$
followed by another $\text{FC}$ layer without activation function, i.e.
$$\text{FFN}(x)=W_2 \cdot ReLU(W_1\cdot x+b_1)+b_2$$
where $W_1, W_2, b_1, b_2$ are the weight matrices and bias vectors respectively for the $\text{FC}$ layers.

The \textbf{Add \& Norm} layer is a residual connection \cite{1512.03385} followed by a layer normalization layers \cite{1607.06450}.
That is,
$$\text{Add \& Norm}(x)=\text{LayerNorm}(x+\text{Sublayer}(x))$$
where $\text{Sublayer}(x)$ is either the multi-head self-attention layer or the feed-forward layer (refer to Figure \ref{fig:attention_is_all_you_need}).

Residual connections (i.e., adding the input to the output of the layer),
also known as skip connections, are commonly used to avoid the vanishing gradient problem \cite{1512.03385}.
In this case, the residual connection is used to avoid the degradation of the model performance
when the model is deep (i.e., when the number of sequential layers $N$ is large).
While we do not use very deep models in this work, we still use residual connections
as in the original architecture for more stable training (refer to Section~\ref{sec:training}).
Moreover, the residual connection is used to propagate the positional information to the next layers
as the attention mechanism is permutation invariant (refer to Section~\ref{sec:attention_mechanism}).

The layer normalization layer \cite{1607.06450}, $\text{LayerNorm}$ is used to normalize the output of the residual connection. The normalization is used to stabilize the training process (however, it is not strictly necessary, refer to Section~\ref{sec:training} for our empirical findings).

This constitutes the main building block of the Transformer encoder architecture. The next section will describe the attention mechanism in detail.

\subsection{Attention mechanism} \label{sec:attention_mechanism}

This section will describe the attention mechanism, its variations and the intuition behind it.
Moreover, we will compare different attention mechanism implementations in terms of
their computational complexity which is important for fast calculations.
Most importantly, we will describe the \textbf{multi-head attention} mechanism which is used in the vanilla Transformer architecture
as a continuation of Section~\ref{sec:general_architecture}.

\subsubsection{Dot-Product Attention and Multi-Head Attention} \label{sec:dot_product_attention}

The attention mechanism introduced in the Transformer architecture \cite{1706.03762} used a \textbf{scaled dot-product attention}.

The main idea of the \textbf{dot-product attention} mechanism is to compute the mapping of a query $q_i$ for each input vector $x_i$ to a set of key-value pairs $(k_j, v_j)$.
The query $q_i$, key $k_i$ and value $v_i$ vectors are simply linear transformations of the input vectors $x_i$,
i.e., $q_i=W_Q\cdot x_i$, $k_i=W_K\cdot x_i$, $v_i=W_V\cdot x_i$ where $W_Q$, $W_K$ and $W_V$ are the weight matrices.
The attention mechanism is a weighted sum of the values $v_j$ where the weights are computed as a function of the query $q_i$ and the key $k_j$.
That is, $Attention(x_i)=\sum_j \alpha_{ij} v_j$ where $\alpha_{ij}=\text{softmax}(q_i \cdot v_i)$ is the weight of the $j$-th value $v_j$.
In practice, the attention mechanism is computed for all the queries $q_i$ at the same time by utilizing the following expression in matrix form:
\begin{equation}
    \begin{array}{rll}
        Q & = W_Q \cdot X, \\
        K & = W_K \cdot X, \\
        V & = W_V \cdot X
    \end{array}
\end{equation}
\begin{equation}\label{eq:dot_product_attention_matrix}
    \text{Attention}(Q, K, V)=\text{softmax}(Q K^T \cdot \frac{1}{\sqrt{d_k}}) V
\end{equation}

Figure \ref{fig:attention_is_all_you_need} visualizes the attention mechanism introduced in \cite{1706.03762}.

\begin{remark}
    In \cite{1706.03762}, authors additionally scaled the weights $\alpha_i$ by the square root of the dimension of the key vectors $d_k$.
    This is, however, not strictly necessary and is done for numerical stability reasons.
\end{remark}

The \textbf{multi-head attention} mechanism is simply a concatenation of multiple attention mechanisms.
That is, we can compute $h$ different attention mechanisms in parallel and then concatenate the results.
The main idea behind this is that different attention mechanisms can learn different features of the input vectors.
After the concatenation, we apply a linear transformation (i.e., a fully-connected layer) to the concatenated vectors to get the final output
with a desired dimensionality which is usually the same as the input vectors $x_i$.

However, in the main paper \cite{1706.03762}, authors implemented the multi-head attention mechanism
in a more efficient way compared to the naive concatenation of multiple attention heads
of size $d_k$. They, instead, linearly transformed the input vectors
to the dimension of $d_k\/h$ where $h$ is the number of attention heads and applied
the attention mechanism to the transformed vectors. This is more efficient because
we require less parameters and can benefit from the increased accuracy of having
multiple attention heads.

\begin{figure}[h!]
    \centering
    \includegraphics[scale=0.15]{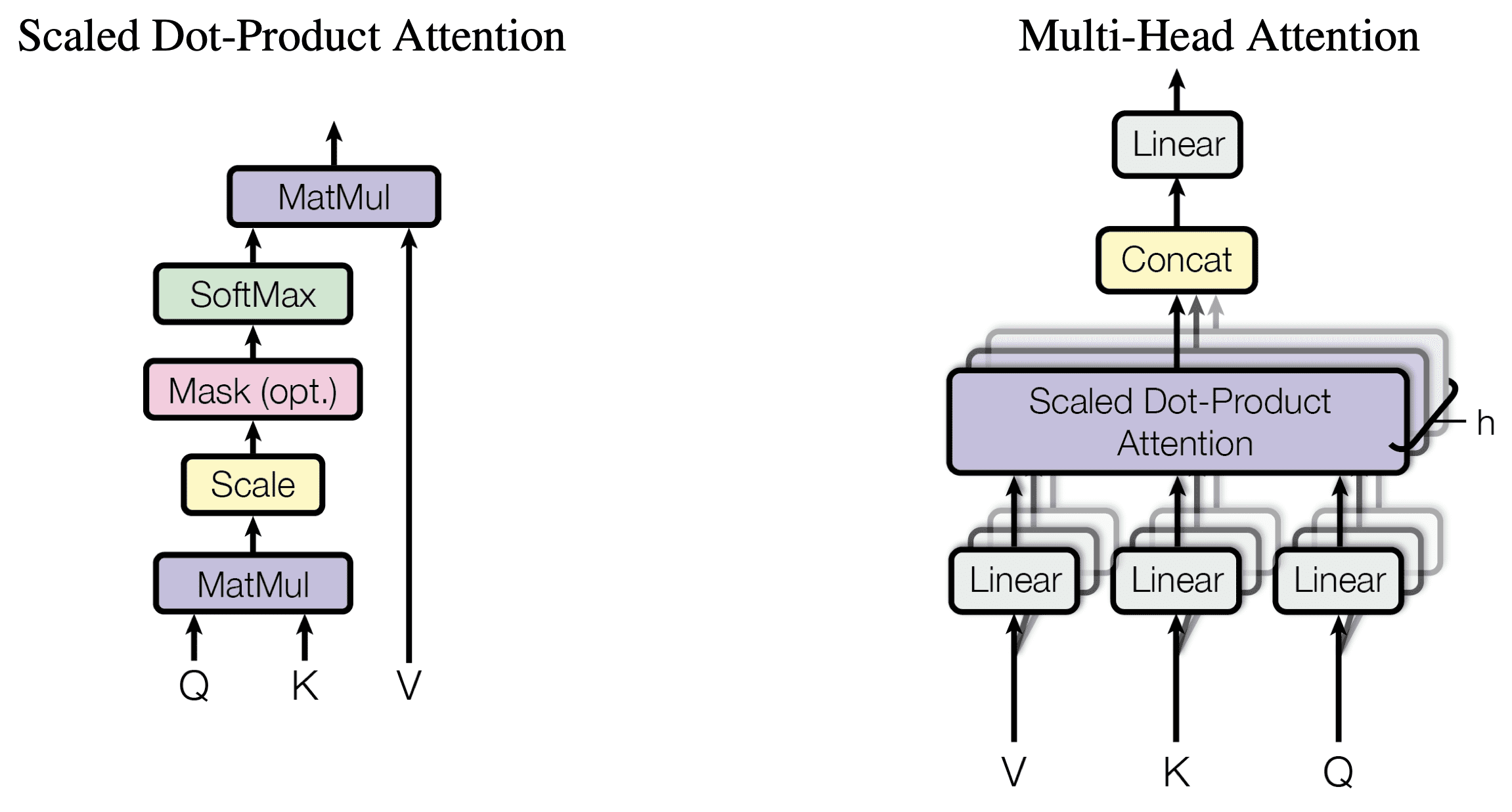}
    \caption{Scaled Dot-Product Attention and Multi-Head Attention \cite{1706.03762}}
    \label{fig:scaled_dot_product_attention}
\end{figure}

\subsubsection{Linear attention} \label{sec:linear_attention}

In \cite{2006.16236}, authors propose an extension to the dot-product attention mechanism called \textbf{linear attention}.
This significantly reduces the computational complexity of the attention mechanism by eliminating the need to compute the softmax function.

Notice that in \ref{eq:dot_product_attention_matrix}, the softmax function is applied rowwise to the matrix $Q K^T$.
The softmax function can be substituted with a general similarity function $\text{sim}(\cdot, \cdot)$ between a query $q_i$ and a key $k_j$.
The equation \ref{eq:dot_product_attention_matrix} for output value $v'_i$ can then be rewritten as follows:
\begin{equation}
    v'_i=\text{Attention}(Q, K, V)_i=\frac{\sum_j \text{sim}(q_i, k_j) v_j}{\sum_j \text{sim}(q_i, k_j)}
\end{equation}

The only constrained imposed on the similarity function $\text{sim}(\cdot, \cdot)$ is that it should be non-negative for it to define an attention function.
This conveniently includes all kernels. That is $\text{sim}(q_i, k_j)=\phi(q_i)^T \phi(k_j)$ where $\phi(\cdot)$ is a feature map.

So that given a kernel with a feature map $\phi(\cdot)$, the attention mechanism can be computed as follows:
\begin{equation}
    v'_i=\text{Attention}(Q, K, V)_i=\frac{\sum_j \phi(q_i)^T \phi(k_j) v_j}{\sum_j \phi(q_i)^T \phi(k_j)}
\end{equation}

And we can rewrite the attention mechanism in matrix form as follows:
\begin{equation}
    \text{Attention}(Q, K, V)=\frac{\phi(Q)^T \phi(K) V}{\phi(Q)^T \phi(K)}
\end{equation}

Regrouping the terms, we get the following expression for the attention mechanism:
\begin{equation}
    \text{Attention}(Q, K, V)=\phi(Q)^T \frac{\phi(K) V}{\phi(Q)^T \phi(K)}
\end{equation}

which makes it evident that we can compute $\sum_j \phi(k_j) v_j$ once and reuse them for all the queries $q_i$
which reduces the computational complexity from $O(N^2)$ to $O(N)$ where $N$ is the number of input vectors
in the attention layer.

\begin{remark}
    In \cite{2006.16236}, authors used the $\phi(x)=elu(x)+1$ feature map where
    $elu(x)=\max(0, x)+\min(0, \alpha(\exp(x)-1))$ is the exponential linear unit activation function.
    This feature map is used to ensure that the attention mechanism is non-negative
    and hence defines a valid attention function. Moreover, $elu(\cdot)$ is used instead of $ReLU(\cdot)$
    to ensure the differentiability when x is negative.
\end{remark}

\subsection{Transformers for time series modelling}

Originally, the Transformer architecture was introduced for Natural Language Processing tasks \cite{1706.03762}
but has since found applications in other domains and time-sereis modelling specifically \cite{2202.07125}.
The motivation behind using the Transformer architecture for time series modelling is that
they show good performance in many sequence modelling tasks. While many
specialized Transformer architectures were proposed for time series modelling
(e.g., \cite{tuli2022tranad}, \cite{2205.13504}) and specifically deployed in
asset management setting \cite{Lezmi2023TimeSF},
in the current work, we will use the least tuned architecture, i.e., the vanilla Transformer architecture
and the architecture with linear attention \cite{2006.16236}. The reason for
using the not so specialized architectures is that we want to show that the Transformer
can achieve good out of the box performance on the anomaly detection task with minimal
tuning and that it can be efficiently implemented on an FPGA board.

\section{FPGA design} \label{sec:fpga_design}

In this section, the main concepts of programming an FPGA will be introduced
and the specific optimizations that can be applied to speed up the
computations.

Readers will be introduced to the common optimization techniques and how they
are achieved. An matrix multiplication example will be provided to illustrate the concepts and
how the optimizations affect the latency and resource utilization.

Lastly, the analysis of the specific optimizations that can be applied to the Transformer architecture
will be provided.

\subsection{Introduction to FPGA programming and development} \label{sec:fpga_development}

In this section, we will introduce the main concepts of FPGA programming and development tools.


\subsubsection{Common Terms}

In this section, common terms will be introduced.
The terms will be used throughout Section~\ref{sec:fpga_design}.
It is not required to read all of them at once, but it is recommended to refer to this section
when a term is not clear (i.e., only when necessary).

\begin{figure}[h!]
    \centering
    \includegraphics[scale=0.7]{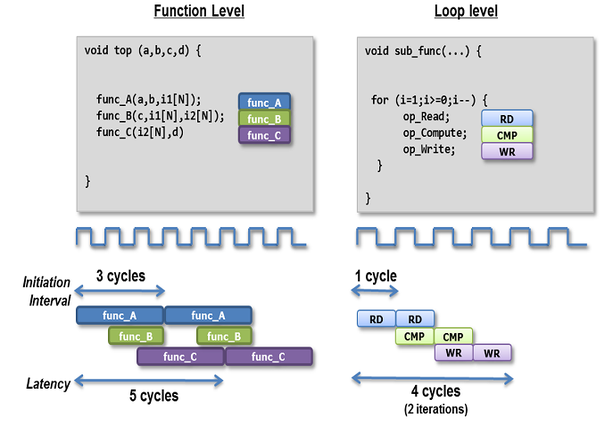}
    \caption{Latency vs Initiation Interval illustration. Source: \cite{VitisYork}}
    \label{fig:latency_vs_ii}
\end{figure}

\begin{definition}
    \textbf{LUT (Look-Up Table)} is a small, fast memory that stores the output of a Boolean function of its inputs.
    The LUT is the basic building block of an FPGA and is capable of implementing any logic function of N Boolean variables.
\end{definition}
\begin{definition}
    \textbf{BRAM (Block RAM)} is a dedicated two-port memory that can be used to store data.
\end{definition}
\begin{definition}
    \textbf{DSP (Digital Signal Processing)} is a specialized hardware unit that is optimized for performing mathematical operations.
\end{definition}
\begin{definition}
    \textbf{Clock cycle} is the time between two consecutive rising edges of the clock signal.
    It is the amount of time between two pulses of an oscillator, a single increment of the
    central processing unit (CPU) clock during which the smallest unit of processor activity
    is carried out.
\end{definition}
\begin{definition}
    \textbf{Latency} is the time between the start of an operation and the moment its results become available
    or the number of clock cycles required to complete an operation.
    \textbf{Latency of a loop} is the number of clock cycles required to complete one iteration of the loop.
\end{definition}
\begin{definition}
    \textbf{Throughput} is the number of operations that can be completed in a given amount of time.
\end{definition}
\begin{definition}
    \textbf{Initiation Interval (II)} is the number of clock cycles between the start of two consecutive iterations of a loop.
    That is, it is the maximum rate (in clock cycles) at which loop iterations can be initiated.
    In the ideal case, the II is equal to 1 so that we can start a new iteration of the loop every clock cycle.
    Initiation interval is different from latency.
    The reason for this is pipelining which will be described in Section~\ref{sec:loop_pipelining}.
    For a visual explanation, see Figure \ref{fig:latency_vs_ii}.
\end{definition}
\begin{definition}
    \textbf{Trip count} is simply the number of iterations of a loop.
\end{definition}

\paragraph{HLS synthesis} \label{sec:hls_synthesis}

In this section, HLS synthesis will be described ~\cite{AMD2023VitisHLS}.
It is now the common workflow in the FPGA development because it significantly
improves the productivity when working with design.

HLS synthesis is a methodology that bridges the gap between high-level programming
languages, such as C, C++, or OpenCL, and the low-level hardware description languages
typically used in FPGA design, like Verilog or VHDL. This approach enables developers
to describe their algorithms and functionality at a higher level of abstraction,
allowing them to focus on the problem-solving aspect rather than the intricacies of hardware implementation.

The HLS process starts with a high-level description of the desired functionality.
This description can be written in familiar programming languages (in this case, C++ HLS),
taking advantage of their abstractions and concise syntax.
The HLS tool then performs a series of transformations on this high-level description to generate
an optimized hardware implementation that can be deployed on an FPGA.
The high-level description is transformed by the HLS tool into a RTL (Register Transfer Level) representation,
which is the low-level hardware description that defines the behavior of the FPGA.

\paragraph{Simulation, Cosimulation}

In this section, the processes of \textbf{simulation} and \textbf{cosimulation} will be described.
In the context of developing for Field-Programmable Gate Arrays (FPGAs),
simulation and cosimulation are two crucial techniques for verifying and testing the
functionality of your design before actually programming it onto the FPGA hardware.

\begin{definition}
    \textbf{Simulation} is the process of running a software-based model of your FPGA design on
    a computer to simulate its behavior.
    The process is very similar to running a software program on a computer for
    the purpose of unit testing certain parts of functionality of your code
    \cite{AMD2023VitisSimCosim}. That is, the simulation does not
    involve any RTL code and is simply a software simulation of the high-level description.
\end{definition}

\begin{remark}
    Simulation might not always capture all aspects of hardware behavior, such as timing delays, which can be critical on FPGAs.
\end{remark}

\begin{definition}
    \textbf{Cosimulation} is a technique that combines simulation of the high-level description with simulation of the generated RTL description.
    This means that the simulations of both the original high-level code and the RTL representation in parallel, comparing their behavior.
    The purpose of cosimulation is to ensure that the high-level synthesis tool accurately transformed the high-level description into the desired RTL behavior.
    \cite{AMD2023VitisSimCosim}.
\end{definition}

\subsubsection{Common optimizations} \label{sec:common_optimizations}

In this section, common optimization techniques and how they are achieved will be introduced.

It is quite common to process data blocks
(for example, a sequence of samples in anomaly detection) using for loops.
For loops are usually the main bottleneck in the performance of the design
and it is the area where most of the optimizations are applied first \cite{AMD2023VitisHLS}.

\paragraph{Loop Pipelining} \label{sec:loop_pipelining}

\begin{figure}[h!]
    \centering
    \includegraphics[scale=0.7]{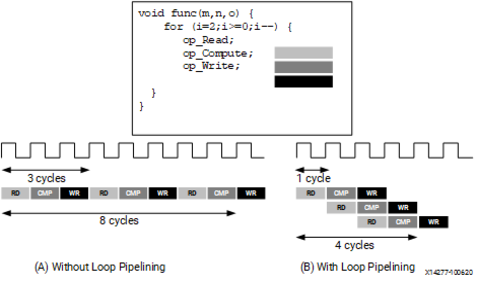}
    \caption{Loop pipelining illustration. Source: \cite{AMD2023Pipelining}}
    \label{fig:loop_pipelining}
\end{figure}

Loop pipelining is a technique used in FPGAs programming to optimize the performance of sequential operations within a loop.
It improves the throughput of loops by breaking them down into multiple stages
that can execute concurrently.
That is, it allows to start the next iteration of a loop before the current
iteration has finished \cite{AMD2023Pipelining}.
Refer to Figure \ref{fig:loop_pipelining} and Figure \ref{fig:latency_vs_ii} for a visual illustration.


Consider example of pipelining a simple for loop which adds 2 vectors (\cite{AMD2023Pipelining}):
\begin{lstlisting}[language=c++,numbers=none]
void toplevel(int* a, int* b, int* c, int len) {
	vadd: for(int i = 0; i < len; i++) {
#pragma HLS PIPELINE
	   c[i] = a[i] + b[i];
	}
}
\end{lstlisting}

Taking the reference numbers from \cite{AMD2023Pipelining},
assume that len is 20 and that one loop iteration takes 3 clock cycles,
then the total latency of the loop is 60 clock cycles without pipelining.
The pipelining pragma \texttt{\#pragma HLS PIPELINE} allows to start
the next iteration of the loop before the current iteration has finished.
By default, the pipelining pragma will try to achieve an II of 1 but this
can be specified manually by using the \texttt{II} parameter.
So, the pipelining pragma reduces the latency of the loop to 22 clock cycles.

In general, the latency of a loop with pipelining is given by the following formula \cite{intelpipelineloopsII}:
\begin{equation}
    \text{Latency} = \text{II} \cdot (\text{Trip Count} - 1) + \text{Loop Body Latency}
\end{equation}


\begin{remark}
    It is not always possible to achieve an II of 1.
    This is because of the dependencies between the iterations of the loop.
    For example, if the loop body depends on the result of the previous iteration,
    then the II cannot be 1 and we have to wait until the previous iteration has finished
    before starting the next one.
\end{remark}


\paragraph{Loop Unrolling}

Loop unrolling is an optimization technique that involves expanding or unwinding
loops in code to potentially enable better utilization of hardware resources and/or
to minimize control flow (branching) in loop iterations.
This technique is not limited to FPGA development but can be particularly useful
in optimizing code for FPGA implementations.

Loop unrolling works by duplicating loop iterations (read as \texttt{copy-pasting}).
This allows utilizing more hardware as the loop body is duplicated multiple times
and loop iterations will utilize different hardware resources.
This increase in performance (i.e., throughput) comes at the cost of increased
resource utilization \cite{AMD2023Pipelining}.

Consider a simple example of a function which multiplies the
input vector of length 4 by a constant, 2:
\begin{lstlisting}[language=c++,numbers=none]
void toplevel(int* a, int* b) {
	smult: for(int i = 0; i < 4; i++) {
#pragma UNROLL
	   b[i] = 2 * a[i];
	}
}
\end{lstlisting}

The unroll pragma \texttt{\#pragma UNROLL} allows to unroll the loop and execute
the loop iterations in parallel.

The unrolled version of the loop will be equivalent to the following code:
\begin{lstlisting}[language=c++,numbers=none]
void toplevel(int* a, int* b) {
	b[0] = 2 * a[0];
	b[1] = 2 * a[1];
	b[2] = 2 * a[2];
	b[3] = 2 * a[3];
}
\end{lstlisting}

\begin{remark}
    It might not be possible to unroll a loop.
    For example, if the trip count is not known at compile time then the loop cannot be unrolled.
    Sometimes it is possible to unroll a loop partially. The unroll pragma allows to specify
    the factor of unrolling, i.e., how many iterations to unroll.
    For example, \texttt{\#pragma UNROLL factor=2} will only duplicate the loop body
    so that there are 2 of them.
\end{remark}

\paragraph{Loop Reordering}

Loop reordering optimization is an optimization used to improve the performance
by changing the order in which loops are executed.
This optimization is not strictly related to FPGA development and
it involves altering the nesting order of loops in a way that improves data locality,
cache utilization (for example, on modern CPU) and enables usage of SIMD resources.

In the context of FPGA, loop reordering can be used achieve better \texttt{II}
when we are dealing with pipelining and nested loops (see Section~\ref{sec:loop_pipelining} and Section~\ref{sec:mm_loop_reordering} for an example).

\paragraph{Function Inlining}

Function inlining optimization technique is not specific to FPGA development only.
The technique is used to improve the performance of a program by reducing the overhead associated with function calls.
Calling a function incurs some overhead in terms of memory and execution time due to the need to set up the function call stack, pass arguments, and jump to the function's code.
Function inlining aims to eliminate this overhead by replacing a function call with the actual body of the function at the call site.
In other words, the compiler takes the contents of the called function and inserts it directly into the location where the function is called.
This, however, increases the size of the code and, specifically in the case of FPGAs, the resource utilization (LUTs and FF).

\paragraph{Array Partitioning and Reshaping}

Partitioning arrays in an FPGA (Field-Programmable Gate Array) refers to the
process of dividing a large memory block (e.g., one array) into smaller sections, often referred to
as memory banks or partitions.

Partitioning allows accessing different parts of the array in parallel
so that bottlenecks caused by a single memory interface being overwhelmed with requests
can be avoided.

\begin{figure}[h!]
    \centering
    \includegraphics[scale=0.5]{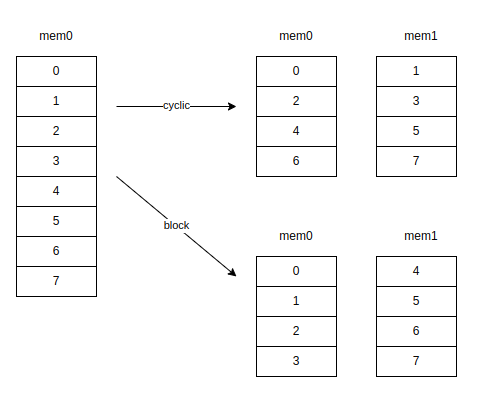}
    \caption{Array partitioning illustration.}
    \label{fig:array_partitioning}
\end{figure}

There are 3 different types of partitioning that can be applied to an array (Refer to Figure \ref{fig:array_partitioning}):

\begin{enumerate}
    \item \textbf{Cyclic}.
          In a cyclic partition, the array is divided blocks of interleaved elements of the original array.
    \item \textbf{Block}.
          In a block partition, the array is divided into non-overlapping blocks of sequential elements in the original array.
          Each block is assigned to a separate memory bank.
    \item \textbf{Complete}.
          The array is split intio intividual elements
\end{enumerate}

The disadvantage of partitioning is that it increases the number of memory interfaces
which leads to the increased resource utilization (i.e., more FFs, LUTs are used because each memory block requires separate control logic).
This can be partly mitigated by using the \textbf{array reshaping}.
The difference between partitioning and reshaping is that partitioning
creates multiple memory interfaces while reshaping still uses a single memory interface
(i.e., all partitions are merged to a single physical memory).

\subsubsection{Example of optimizing matrix multiplication}

In this section, we will describe the process of optimizing a matrix multiplication
using the techniques described in Section~\ref{sec:common_optimizations}.
This section can be treated as a tutorial on how to optimize a simple matrix multiplication.

Full source code for with all the files can be located in \texttt{vitis\_hls/matmul\_naive}.

\paragraph{Naive implementation}

The naive implementation of matrix multiplication is simply a triple for loop
which directly implements the definition of $C = A \cdot B$ where A, B and C are the matrices
and $C_{i,j}$ are defined as in Equation \ref{eq:matmul}.

\begin{equation} \label{eq:matmul}
    C_{i,j} = \sum_{k=0}^{N} A_{i,k} \cdot B_{k,j}
\end{equation}

\begin{lstlisting}[language=c++,numbers=none]
#include "matrixmul.h"

void matmul(mat_a_t a[MAT_A_ROWS][MAT_A_COLS],
            mat_b_t b[MAT_B_ROWS][MAT_B_COLS],
            result_t res[MAT_A_ROWS][MAT_B_COLS]) {
loop_i:
  for (int i = 0; i < MAT_A_ROWS; i++) {
  loop_j:
    for (int j = 0; j < MAT_B_COLS; j++) {
      res[i][j] = 0;
    loop_k:
      for (int k = 0; k < MAT_B_ROWS; k++) {
#pragma HLS PIPELINE off
        res[i][j] += a[i][k] * b[k][j];
      }
    }
  }
}
\end{lstlisting}

\begin{figure}[h!]
    \centering
    \includegraphics[width=\textwidth]{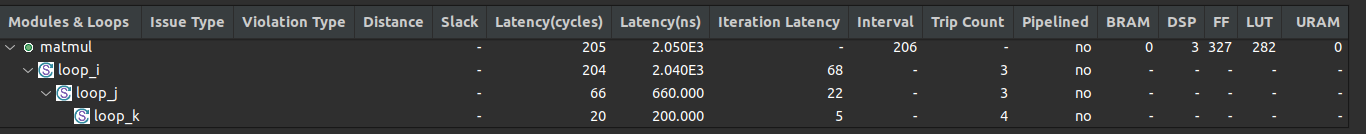}
    \caption{Naive matrix multiplication synthesis report}
    \label{fig:matmul_naive_synth}
\end{figure}

The Vitis HLS Synthesis Report for the naive implementation is presented in Figure \ref{fig:matmul_naive_synth}.
The data types for matrices used are int32\_t. Matrix A is of size 3x4 and matrix B has size 4x3.
For this baseline implementation, we are reaching a latency of 205 clock cycles, using 327 FF and 282 LUTs.

\paragraph{Loop pipelining and unrolling} \label{sec:loop_pipelining_unrolling}

The baseline code can be optimized by pipelining.
There are three loops that can be pipelined (\texttt{loop\_i}, \texttt{loop\_j} and \texttt{loop\_k}).

\textbf{Case 1: Pipelining \texttt{loop\_k}.} A simple pipelining of the innermost loop
leads to the pipelining of Multiply and Accumulate operation (MAC) which is the main operation in the loop.
There is no need to partition arrays a and b as memory in arrays a and b only need
to supply 1 element per cycle.

The pipelining leads to decrease in latency to 42 clock cycles at the cost of using
526 FF and 501 LUT which is almost twice as many resources as in the naive implementation.

\begin{figure}[h!]
    \centering
    \includegraphics[width=\textwidth]{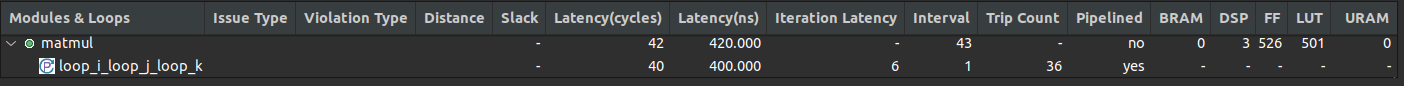}
    \caption{Matrix multiplication with pipelined loop \texttt{loop\_k}}
    \label{fig:matmul_pipek_synth}
\end{figure}

\textbf{Case 2: Pipelining \texttt{loop\_j}.}
The pipelining of the middle loop requires partitioning of arrays a by \texttt{MAT\_A\_COLS} and b by \texttt{MAT\_B\_ROWS}.
There are \texttt{MAT\_A\_COLS} (or \texttt{MAT\_B\_ROWS}) MAC operations per cycle so that
memory of a and b needs to be sufficiently divided to supply \texttt{MAT\_B\_ROWS} elements per cycle.
Array a is partitioned along the second dimension and array b is partitioned along the first dimension
because of the access patterns.
In this example, we use the \texttt{complete} partitioning which divides the memory into
individual registers.

\begin{lstlisting}[language=c++,numbers=none]
#include "matrixmul.h"

void matmul(mat_a_t a[MAT_A_ROWS][MAT_A_COLS],
            mat_b_t b[MAT_B_ROWS][MAT_B_COLS],
            result_t res[MAT_A_ROWS][MAT_B_COLS]) {
#pragma HLS ARRAY_PARTITION variable = a complete dim = 2
#pragma HLS ARRAY_PARTITION variable = b complete dim = 1
loop_i:
  for (int i = 0; i < MAT_A_ROWS; i++) {
#pragma HLS PIPELINE off
  loop_j:
    for (int j = 0; j < MAT_B_COLS; j++) {
      int tmp = 0;
#pragma HLS PIPELINE
    loop_k:
      for (int k = 0; k < MAT_B_ROWS; k++) {
#pragma HLS UNROLL
        tmp += a[i][k] * b[k][j];
      }
      res[i][j] = tmp;
    }
  }
}
\end{lstlisting}

\begin{figure}[h!]
    \centering
    \includegraphics[width=\textwidth]{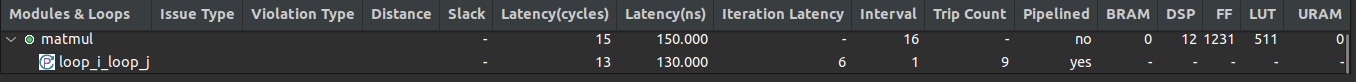}
    \caption{Matrix multiplication with pipelined loop \texttt{loop\_j}}
    \label{fig:matmul_pipej_synth}
\end{figure}

With this addition, the latency is down to 15 clock cycles and the resource usage
has increased to 1221 FF and 511 LUTs (Figure \ref{fig:matmul_pipej_synth}).

\textbf{Case 3: Pipelining \texttt{loop\_i}.} Pipelining the outermost loop requires
additional partitioning of array res by \texttt{MAT\_B\_COLS} and full partitioning
of array b as \texttt{loop\_k} and \texttt{loop\_j} are getting unrolled.

\begin{lstlisting}[language=c++,numbers=none]
#include "matrixmul.h"

void matmul(mat_a_t a[MAT_A_ROWS][MAT_A_COLS],
            mat_b_t b[MAT_B_ROWS][MAT_B_COLS],
            result_t res[MAT_A_ROWS][MAT_B_COLS]) {
#pragma HLS ARRAY_PARTITION variable = a complete dim = 2
#pragma HLS ARRAY_PARTITION variable = b complete dim = 0
#pragma HLS ARRAY_PARTITION variable = res complete dim = 2
loop_i:
  for (int i = 0; i < MAT_A_ROWS; i++) {
#pragma HLS PIPELINE
  loop_j:
    for (int j = 0; j < MAT_B_COLS; j++) {
#pragma HLS UNROLL
      int tmp = 0;
    loop_k:
      for (int k = 0; k < MAT_B_ROWS; k++) {
#pragma HLS UNROLL
        tmp += a[i][k] * b[k][j];
      }
      res[i][j] = tmp;
    }
  }
}
\end{lstlisting}

\begin{figure}[h!]
    \centering
    \includegraphics[width=\textwidth]{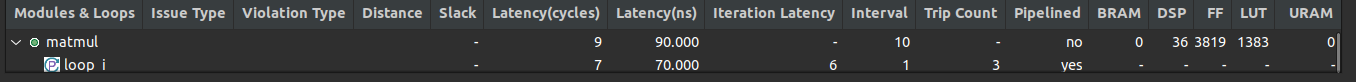}
    \caption{Matrix multiplication with pipelined loop \texttt{loop\_i}}
    \label{fig:matmul_pipei_synth}
\end{figure}

The latency is down to 9 clock cycles and the resource usage is up to 3819 FF and 1383 LUTs
(Figure \ref{fig:matmul_pipei_synth}).

\paragraph{Loop reordering} \label{sec:mm_loop_reordering}

We can also compare how the loop reordering affects the performance of the matrix multiplication
and specifically how it compares to the pipelining of the \texttt{loop\_j} in terms of latency and resource usage.

The loop reordering technique allows us to avoid partitioning matrix a compared to the \texttt{loop\_j} pipelining solution. However,
we now have to partition the output matrix by \texttt{MAT\_B\_COLS}.

\begin{lstlisting}[language=c++,numbers=none]
#include "matrixmul.h"

void matmul(mat_a_t a[MAT_A_ROWS][MAT_A_COLS],
            mat_b_t b[MAT_B_ROWS][MAT_B_COLS],
            result_t res[MAT_A_ROWS][MAT_B_COLS]) {
  int temp_sum[MAT_B_COLS];
#pragma HLS ARRAY_PARTITION variable = b dim = 2 complete
#pragma HLS ARRAY_PARTITION variable = res dim = 2 complete
#pragma HLS ARRAY_PARTITION variable = temp_sum dim = 1 complete
loop_i:
  for (int i = 0; i < MAT_A_ROWS; i++) {
  loop_k:
    for (int k = 0; k < MAT_B_ROWS; k++) {
#pragma HLS PIPELINE
    loop_j:
      for (int j = 0; j < MAT_B_COLS; j++) {
#pragma HLS UNROLL
        int result = (k == 0) ? 0 : temp_sum[j];
        result += a[i][k] * b[k][j];
        temp_sum[j] = result;
        if (k == MAT_B_ROWS - 1) {
          res[i][j] = result;
        }
      }
    }
  }
}
\end{lstlisting}

\begin{figure}[h!]
    \centering
    \includegraphics[width=\textwidth]{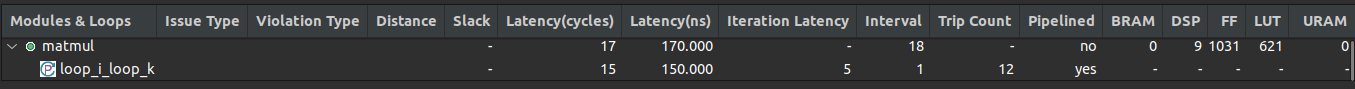}
    \caption{Matrix multiplication with loop reordering}
    \label{fig:matmul_loop_reordering}
\end{figure}

From Figure \ref{fig:matmul_loop_reordering}, the latency is 17 clock cycles
and the resource usage is 1031 FF and 611 LUTs.
The performance is worse than the pipelined loop \texttt{loop\_j} but the resource usage is lower for FF and higher for LUTs.

\textbf{Summary}
The pipelining of the outermost loop leads to the best performance with the lowest latency.
However, this comes at the big cost of using the most resources as array b
needs to be fully partitioned.
In this work, we will use the pipelined loop \texttt{loop\_j}
as it provides a good trade-off between latency and resource usage.

\subsection{Transformer architecture optimizations} \label{sec:transformer_optimizations}

In this section, we present the specific optimizations that were applied to the Transformer architecture.
For HLS code (i.e., implementation), see \texttt{vitis\_hls} directory.
The comparison of the performance and resource utilization of the FPGA optimized architecture
and the CPU implementation is presented in Section~\ref{sec:performance}.

\subsubsection{Vanilla Transformer} \label{sec:fpga_vanilla_transformer}

In this section, we will describe the specific optimizations that can be applied to the Transformer architectures
and present the HLS synthesis reports for the optimized architecture.

For the purpose of the analysis, we will use the architectures with the
hyperparameters as described in Section~\ref{sec:architecture}.
The models achieve reasonable performance while still being small enough
to fit the board.

For all \textbf{matrix multiplications} performed in the architecture, we will use the pipelined loop \texttt{loop\_j}
as described in Section~\ref{sec:loop_pipelining_unrolling} because
it provides a reasonable performance-resource utilization tradeoff.

We utilize \texttt{float} \textbf{data type} for all the matrices in the architecture.
We found that using \texttt{double} data type does not affect the accuracy performance
of the model, while it significantly increases the resource utilization (DSP)
and makes the model not fit into the board.

We also use the following numerical optimization technique applied to \textbf{softmax}
to make it numerically stable.
Notice that
\begin{align*}
    \text{softmax}(x_i) & = \frac{e^{x_i}}{\sum_{j=1}^{N} e^{x_j}}                     \\
                        & = \frac{e^{x_i - \max(x)}}{\sum_{j=1}^{N} e^{x_j - \max(x)}} \\
                        & = \text{softmax}(x_i - \max(x))
\end{align*}
Since $e^x_i$ term can be very large, subtracting the maximum value from all the elements
bounds the values of $e^x_i$ (the largest element will be 0) so that the
sum of exponents will never become large.

For the \textbf{ReLU} activation function, we which needs to be applied to
a 2 dimensional matrix (see Feed Forward layer of the encoder block in Section~\ref{sec:encoder_block})
fully unroll the loops and use the \texttt{complete} partitioning of the matrix.

All the loops in the architecture are automatically pipelined unless
it is impossible to do so (e.g., because of the dependencies between the iterations).

The optimized architecture reaches latency of 3714 cycles (37.14 microseconds) utilizes 5\% BRAM, 30\% FFs and 90\% LUTs.
\newline
See full report in \texttt{vitis\_hls/transformer/csynth.rpt}.

For comparison, the architecture with all optimization turned off
reaches latency of 34745 cycles (347.45 microseconds, 10 times as many) and utilizes only 2\% BRAM, 2\% FFs and 10\% LUTs.
\newline
See full report in \texttt{vitis\_hls/transformer\_slow/csynth.rpt}.

\subsubsection{Linear transformer}

We also implement the linear transformer architecture (i.e., linear attention)
as described in Section~\ref{sec:linear_attention}
on the board.

Similarly to the vanilla transformer implementation (described in Section~\ref{sec:fpga_vanilla_transformer}),
we also utilize \texttt{float} data type for all the matrices in the architecture,
use \texttt{loop\_j} pipelining for matrices, fully unroll the loops for ReLU activation function.

\begin{remark}
    As an implementation detail, the matrices had to be manually partitioned
    with the usage of \texttt{pragma HLS ARRAY\_PARTITION} directive because
    Vitis HLS could not infer the partitioning automatically.
\end{remark}

The optimized version achieves 2986 cycles (29.86 microseconds) latency and utilizes 18\% BRAM, 43\% FFs and 99\% LUTs
\newline
The full report is in \texttt{vitis\_hls/linear\_transformer/csynth.rpt}.

For comparison, the architecture with all optimization turned off reaches latency of 68894 cycles (688.94 microseconds) and utilizes 17\% BRAM, 6\% FFs and 24\% LUTs.
The full report is in \texttt{vitis\_hls/linear\_transformer\_slow/csynth.rpt}.

\section{Experiments} \label{sec:experiments}

In this section, we will describe the experiments that were performed to evaluate the performance of the proposed architecture.
While the main focus of this work is the inference performance of the models (refer to Section~\ref{sec:performance}),
we will also describe the training procedure and the accuracy performance of the models.

\subsection{Architecture and hyperparameters} \label{sec:architecture}

Here we will describe the model architecture and the hyperparameters used for the experiments.

In the experiments 3 models were used for comparison:
\begin{itemize}
    \item \textbf{Linear Regression} - a simple linear regression model on handcrafted features
    \item \textbf{Transformer Encoder} - a transformer encoder model on raw time series data
    \item \textbf{Linear Transformer Encoder} - a linear transformer model on raw time series data
\end{itemize}

The reason for using the linear regression model is to provide a baseline for the performance of the transformer models
to show that even simple untuned models can outperform the linear regression model.

For both transformer models, we used the encoder architectures as described in Section~\ref{sec:attention_mechanism}
with a linear layer on top of the output of the transformer encoder to get the final prediction (i.e., if a sample is an anomaly).
\begin{remark}
    The encoder part has positional encoding and layer normalization disabled.
    We found that the positional encoding does not improve the performance of the model and leads to more
    unstable training (see Section~\ref{sec:training}).
\end{remark}

The transformer hyperparameters used for the experiments are presented in Table \ref{tab:hyperparameters}.

\begin{table}[h!]
    \centering
    \begin{tabular}{|c|c|c|c|c|}
        \hline
        Parameter, Model            & Transformer Encoder & Linear Transformer Encoder \\
        \hline
        Window Size                 & 8                   & 8                          \\
        Number of heads             & 8                   & -                          \\
        Dim. of FeedForward network & 16                  & 16                         \\
        Number of blocks            & 2                   & 1                          \\
        \hline
    \end{tabular}
    \caption{Hyperparameters}
    \label{tab:hyperparameters}
\end{table}

The learning rate was chosen using the learning rate finder \cite{1506.01186} (see Section~\ref{sec:learning_rate_finder})
and the batch size was chosen to be the maximum value that
fitted the dataset in memory ($2^{13}$ samples),
we also used \textbf{Adam} optimizer \cite{Kingma2014AdamAM} (see Section~\ref{sec:optimizer}).

\subsubsection{Model Fitting} \label{sec:training}

In this section, we will describe the training procedure,
the main tools used and issues that we encountered and how they were addressed.

\paragraph{General procedure}

The General process of training a neural network involves iteratively adjusting its
parameters to minimize a specified loss function.
This is typically achieved through an optimization algorithm, such as gradient descent.

That is, at each iteration, the parameters of the model $\theta$ are updated as follows:
\begin{equation} \label{eq:gradient_descent}
    \theta_{t+1}=\theta_t - \alpha \nabla_{\theta} \mathcal{L}(\theta_t)
\end{equation}

where $\alpha$ is the learning rate and $\mathcal{L}(\theta_t)$ is the loss function at the $t$-th iteration.

While the general procedure is simple there are multiple methods to improve the convergence of the optimization algorithm,
which will be described in the following sections.

\paragraph{Optimizer} \label{sec:optimizer}

The optimizer is the algorithm that is used to update the parameters of the model.
The general procedure of updating the parameters as described in Equation \ref{eq:gradient_descent},
can be improved by using more sophisticated optimization algorithms.

The \textbf{Adam} optimizer is a popular optimization algorithm used in training artificial neural networks
and is a reasonable baseline choice for many problems.
It stands for "Adaptive Moment Estimation" and combines ideas from two other
optimization techniques: RMSprop (Root Mean Square Propagation) and Momentum.
The Adam optimizer is known for its efficiency and  fast convergence, and robustness
to various types of neural network architectures and problem domains.

The optimization procedure performed by Adam is as follows:
\begin{enumerate}
    \item Initialize 2 moving average accumulators, $m$ and $v$ with 0.
          They will store the exponentially decaying average of past gradients
          and squared gradients respectively for each parameter $\theta$
    \item At each iteration, compute the gradient of the loss function with
          respect to the parameters, $\nabla_{\theta} \mathcal{L}(\theta_t)$
    \item Update the moving averages $m$ and $v$ as follows:
          \begin{align*}
              m & = \beta_1 m + (1 - \beta_1) \nabla_{\theta} \mathcal{L}(\theta_t)   \\
              v & = \beta_2 v + (1 - \beta_2) \nabla_{\theta} \mathcal{L}(\theta_t)^2
          \end{align*}
          where $\beta_1$ and $\beta_2$ are the hyperparameters that control the decay rate of the moving averages.
          The usual values for $\beta_1$ and $\beta_2$ are 0.9 and 0.999 respectively.
    \item Compute the bias-corrected moving averages $\hat{m}$ and $\hat{v}$ as follows:
          \begin{align*}
              \hat{m} & = \frac{m}{1 - \beta_1^t} \\
              \hat{v} & = \frac{v}{1 - \beta_2^t}
          \end{align*}
          The bias correction is necessary because the moving averages are initialized with 0
          (see original paper for derivations \cite{Kingma2014AdamAM}).
    \item Update the network parameters as follows:
          \begin{equation*}
              \theta_{t+1}=\theta_t - \alpha \frac{\hat{m}}{\sqrt{\hat{v}} + \epsilon}
          \end{equation*}
          where $\epsilon$ is a small constant used for numerical stability
          (i.e., to avoid division by zero).

          The update is similar to the gradient descent update rule (Equation \ref{eq:gradient_descent})
          except that the gradient is divided by the square root of the moving average of squared gradients.
\end{enumerate}

We found that the optimizer with default parameters ($\beta_1$ and $\beta_2$) works well for our task, and we did not experiment
with other optimizers.

\paragraph{Learning rate finder} \label{sec:learning_rate_finder}

The learning rate $\alpha$ is one of the most important hyperparameters that determines the step size in parameter space
during gradient descent optimization. An appropriate learning rate is essential for
model convergence.
The problem of choosing the learning rate is a well-known problem in machine learning
and badly chosen learning rate can lead to either underfitting where the model
learns too slowly or it can lead to divergence where the parameters are updated too abruptly.

In \cite{1506.01186}, authors proposed a simple method to find an appropriate learning rate
automatically by plotting the loss function against the learning rate.

The procedure is performed as follows:
\begin{enumerate}
    \item Start with a very small learning rate $\alpha$ and increase it at each iteration
    \item At each iteration, train the model for a few epochs and compute the loss function
    \item Plot the loss function against the learning rate
          This plot is crucial in identifying the "sweet spot" in the learning rate range
          where the loss is decreasing effectively.
    \item Choose the point on the learning rate vs. loss curve where the
          loss starts to decrease most steeply.  This point indicates that the model
          is making the most significant progress towards convergence.
\end{enumerate}

\begin{remark}
    While there are no guarantees that the learning rate finder will find the optimal learning rate,
    it provides a good empirical estimate of the optimal learning rate.
\end{remark}

Instead of manually plotting the loss function against the learning rate, we used the implementation
provided in Pytorch Lightning \cite{lightning.ai_2023_lr_finder}.

\paragraph{Gradient explosion and Gradient clipping} \label{sec:gradient_clipping}
A different challenge of training neural networks is the phenomenon known as the "gradient explosion problem."
We have found that the gradient explosion problem is especially prevalent in the proposed architectures.

\begin{figure}[h!]
    \centering
    \includegraphics[width=\textwidth]{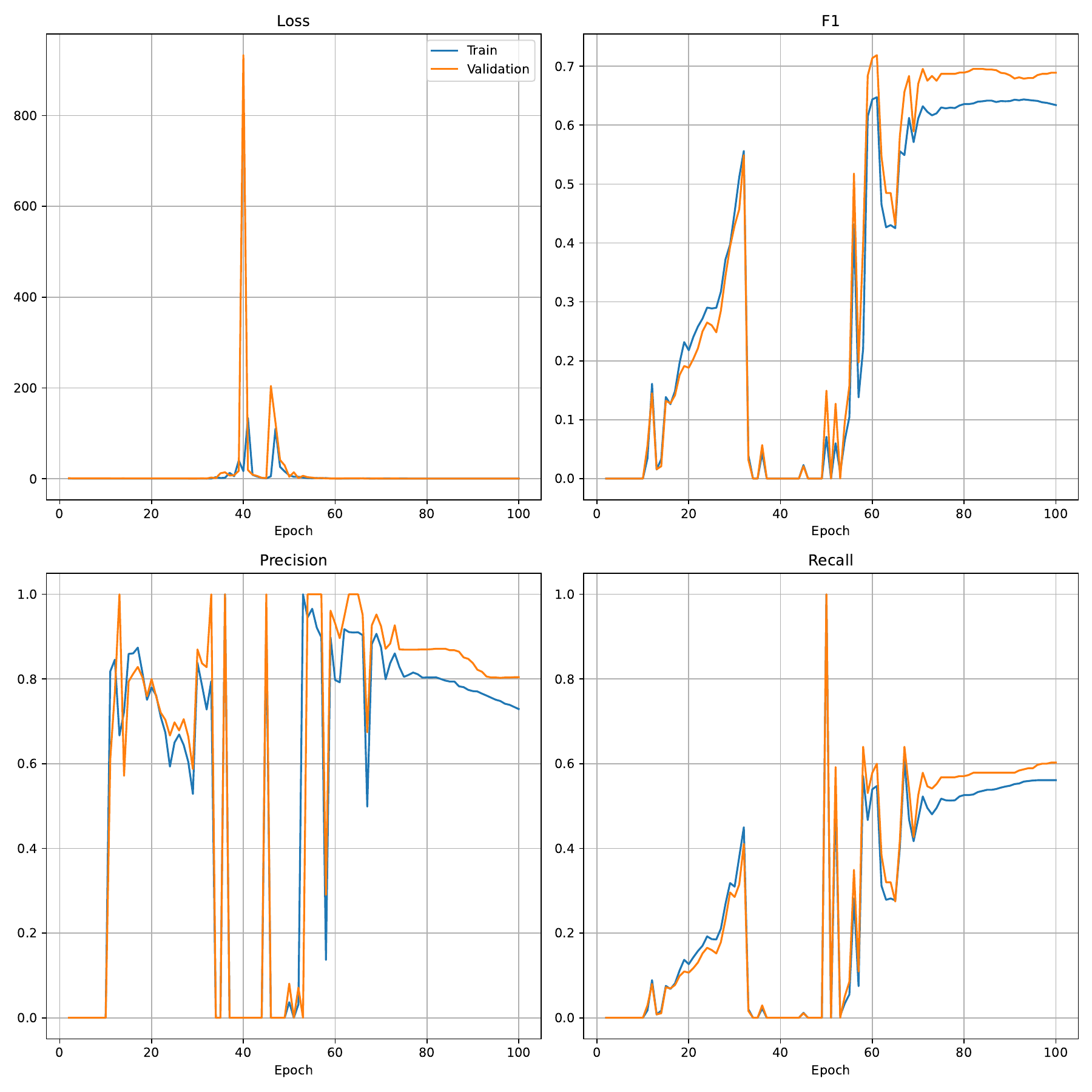}
    \caption{Example of gradient explosion at around epoch 40 which leads to the loss function diverging for a few following epochs.}
    \label{fig:gradient_explosion}
\end{figure}

The gradient explosion problem occurs when the gradient of the loss function with respect to the parameters
becomes too large and the parameters are updated too abruptly (e.g., as in Equation \ref{eq:gradient_descent}).
This can lead to the model diverging and the loss function increasing instead of decreasing.
An example of the loss function diverging is presented in Figure \ref{fig:gradient_explosion}.

To solve the problem of gradient explosion, we used the gradient clipping technique introduced
in \cite{1211.5063}.
The idea behind gradient clipping is to clip the gradient to a maximum value $g_{max}$.
This remediates the problem of gradient explosion because the gradient is bounded and the parameters
are updated more smoothly.

\paragraph{Class imbalance and loss functions} \label{sec:class_imbalance}

In anomaly detection task the dataset is often imbalanced, i.e., the number of normal samples
is much larger than the number of anomalous samples (see Section~\ref{sec:datasets}).

This poses a problem for the training of the model because the model can simply learn to predict
the majority class (i.e., normal samples) and achieve a high accuracy without having
good recall (refer to Section~\ref{sec:metrics} for the description to the metrics).

The loss functions that we use for training is the binary cross-entropy loss function
\cite{Good_1952}
which is computed as follows:
\begin{equation} \label{eq:binary_cross_entropy}
    \mathcal{L}(\theta)=-\frac{1}{N} \sum_{i=1}^N y_i \log(\hat{y}_i) + (1-y_i) \log(1-\hat{y}_i)
\end{equation}
where $y_i$ is the true label and $\hat{y}_i$ is the predicted label.

While the binary cross-entropy loss function is a good choice for the anomaly detection task,
it does not take into account the class imbalance problem.

A way to solve the class imbalance problem is to use a weigh positive
samples (i.e., anomalous samples) more than negative samples in the loss function.

So we can modify the loss function as follows:
\begin{equation} \label{eq:weighted_binary_cross_entropy}
    \mathcal{L}(\theta)=-\frac{1}{N} \sum_{i=1}^N w_i (y_i \log(\hat{y}_i) + (1-y_i) \log(1-\hat{y}_i))
\end{equation}
where $w_i$ is the weight of the $i$-th sample.
In the experiments, we found that weighting the positive samples 5 times more than the negative samples
yielded good results for most of the datasets which we used throughout the experiments.

\paragraph{Training stability: Layer normalization and positional encoding}

While the base transformer encoder architecture uses layer normalization and positional encoding layers,
we found that they lead to unstable training and worse performance of the model on most
of the datasets. Hence, we disabled them for the experiments and replaced them with
identity layers.

\paragraph{Data Preprocessing}

Although, this is not a part of the training procedure, before fitting
a neural network, it is usually a good idea to preprocess the data by scaling it.

For the experiments, we used the \textbf{MinMaxScaler} from the \texttt{scikit-learn} library \cite{scikit-learn},
to scale the input features to the range of $[0, 1]$ for a more stable training.

\subsection{Datasets} \label{sec:datasets}

In this section, the datasets used for model training and performance evaluation will be described.

\subsubsection{Numenta Anomaly Benchmark (NAB)}
To assess the accuracy of predictions, we use the Numenta Anomaly Benchmark ~\cite{Ahmad2017Unsupervised} dataset,
which contains various real-world labeled time series of temperature sensor readings, CPU utilization of cloud machines, service
request latencies, and taxi demands in New York City. It is commonly used to assess the performance of anomaly detection
models on time-series data.

The reason why we use this dataset is that it is a standard benchmark dataset
for anomaly detection in time series and because it has a large number of labeled time series.

\begin{figure}[h!]
    \centering
    \includegraphics[width=\textwidth]{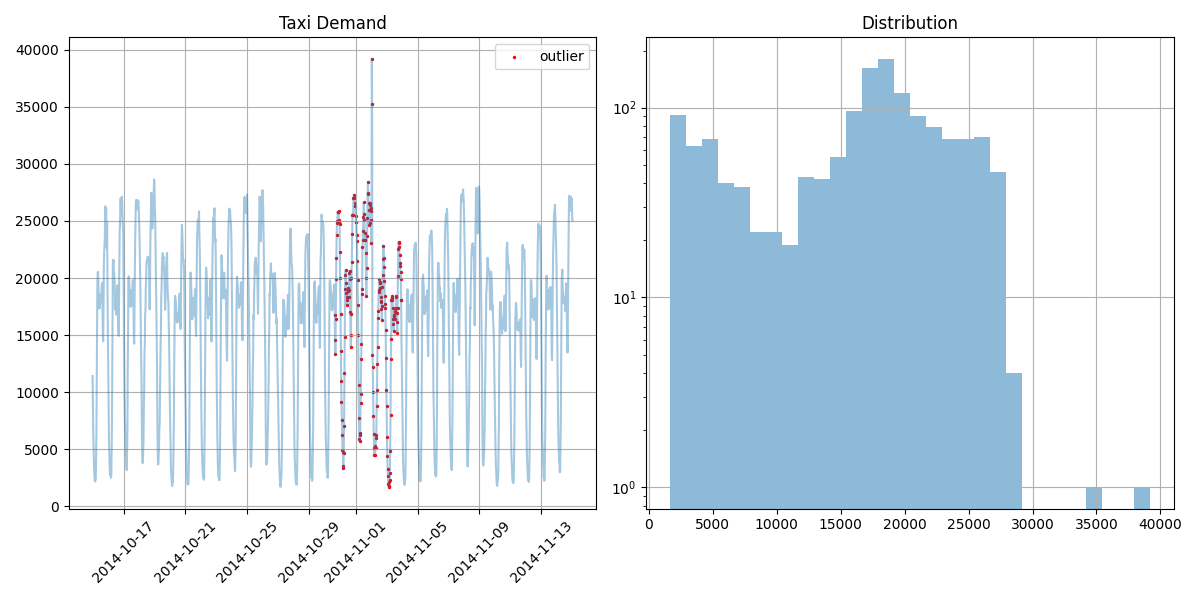}
    \caption{NYC Taxi demand - anomalies highlighted in red. Left: Time series of observations, Right: Distribution of observations.}
    \label{fig:NAB_example_nyc_taxi}
\end{figure}

A sample time series of NYC taxi demand is presented in Figure \ref{fig:NAB_example_nyc_taxi}.

In this work, we only used the NYC taxi demand dataset as in the \cite{tuli2022tranad}.
The dataset consists of 10320 samples with 10\% of anomalous samples.

\subsubsection{KPI Anomaly Detection Dataset}

The other labeled dataset that we use is the KPI Anomaly Detection Dataset (KPI AIOps) \cite{2208.03938}.
This dataset alongside the NAB dataset will be used to evaluate the predictive performance of the anomaly detection models.

The dataset consists of KPI (key performace index) time series data from many
real scenarios of Internet companies with ground truth label. KPIs fall into two
broad categories: service KPIs and machine KPIs. Service KPIs are performance metrics
that reflect the size and quality of a Web service, such as page response time, page views,
and number of connection errors. Machine KPIs are performance indicators that reflect
the health of the machine (server, router, switch), such as CPU utilization, memory utilization,
disk IO and network card throughput.

\begin{figure}[h!]
    \centering
    \includegraphics[width=\textwidth]{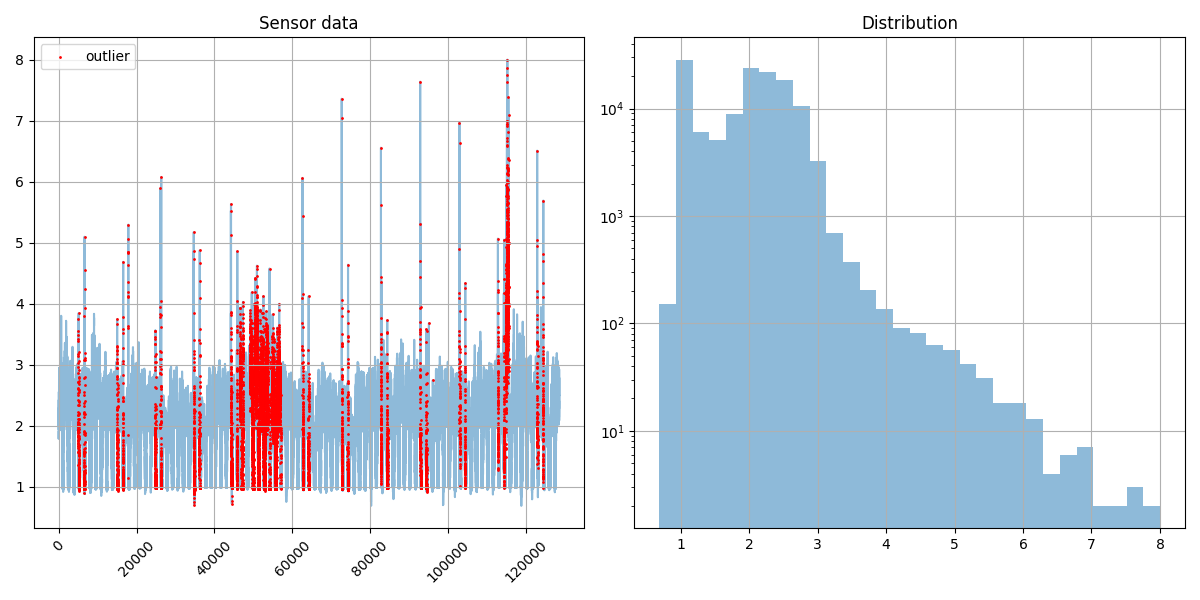}
    \caption{Sensor data from a machine in a data center. The red dots indicate the anomalies. Left: Time series of observations, Right: Distribution of observations.}
    \label{fig:KPIAIOps_example}
\end{figure}

A sample time series of a sensor readings is presented in Figure \ref{fig:KPIAIOps_example}.
We can clearly see the outliers for some of the observations (colored in red).

\subsubsection{FI2010}

In \cite{1705.03233}, authors described the first publicly available benchmark dataset of high-frequency limit order markets for mid-price prediction.
The dataset contains 10-day limit order book data from June 2010 for five stocks that are listed on the Helsinki exchange.
Each entry in the time series provides price details and aggregate order sizes for the top ten levels on both the bid and offer sides of the market,
totaling forty data points. The time series consists of approximately four million messages, representing incoming buy/sell orders or cancellations.
The dataset features order book snapshots taken after every 10 messages, resulting in approximately 400,000 records for the five stocks.

A number of versions of the dataset are available using different normalization schemes. We used the not normalized version of the dataset.

For the purpose of this work, we only extract only the mid price from the dataset which will be used for anomaly detection task.

\paragraph{Synthetic outliers}

Since the dataset is not labeled, we have to inject synthetic anomalies into the dataset.
We employ the approach similar to \cite{Crepey2022Anomaly} with a slight modification.
The algorithm can be summarized as follows:
\begin{enumerate}
    \item Select $n$ samples from the time series which will be contaminated (i.e., anomalous)
    \item Replace the sample $S_i$ with $\hat{S}_i=S_i(1+\delta)$ where $\delta$ is the injected outlier in the return space.
\end{enumerate}

Authors model $\delta$ as a uniformly distributed random variable $\mathcal{U}[0, \rho]$.
We instead use the uniform distribution $\mathcal{U}[-\rho, \rho]$ (i.e., just increase the range)
to allow for positive and negative outliers.

\begin{figure}[h!]
    \centering
    \includegraphics[width=\textwidth]{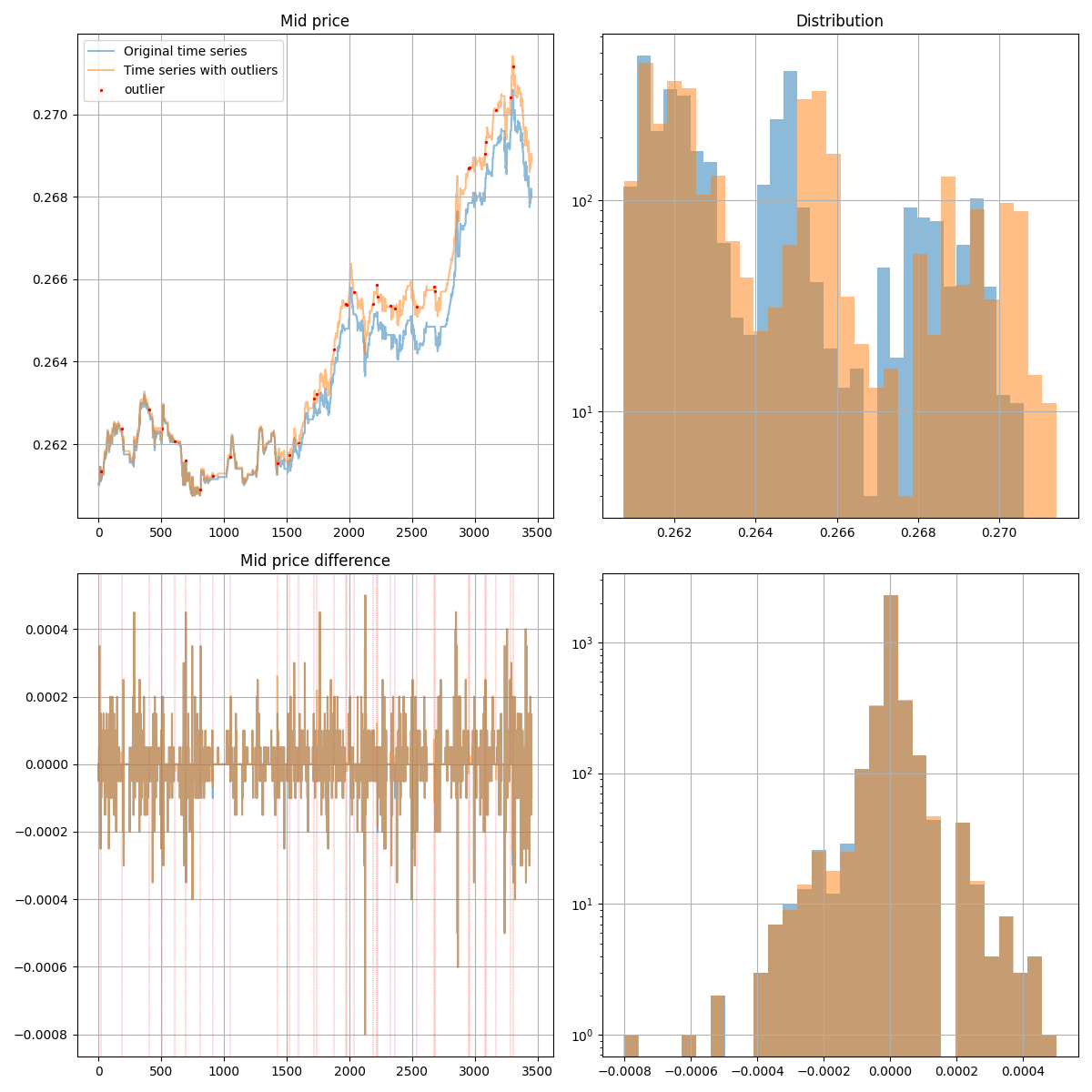}
    \caption{Example of the injected outliers in the FI2010 dataset.}
    \label{fig:FI2010_example_outliers_injected}
\end{figure}

An example of the injected outliers is presented in Figure \ref{fig:FI2010_example_outliers_injected}

\subsection{Accuracy}

In this section, we will describe the main metrics used to evaluate the performance of the anomaly detection models.
The Transformer encoder model will be compared with the simple linear regression model on handcrafted features
and with the Linear Transformer model \cite{2006.16236}.
The inference procedure will be described and the results will be presented.

\subsubsection{Inference}
After training the model on the training set as described in Section~\ref{sec:training}, we can use the model to make predictions on the test set.

\subsubsection{Metrics} \label{sec:metrics}


In this section the metrics used to evaluate the performance of the anomaly detection models will be described.

Before we describe the metrics, we need to introduce the confusion matrix and
the following notation:
\begin{itemize}
    \item \textbf{TP} - True Positive calculated as the number of correctly predicted anomalies
    \item \textbf{TN} - True Negative calculated as the number of correctly predicted non-anomalies
    \item \textbf{FP} - False Positive which is the number of incorrectly predicted anomalies
    \item \textbf{FN} - False Negative which is the number of incorrectly predicted non-anomalies
    \item \textbf{P} - Number of positive samples, i.e., $\textbf{P}=\textbf{TP}+\textbf{FN}$
    \item \textbf{N} - Number of negative samples, i.e., $\textbf{N}=\textbf{TN}+\textbf{FP}$
\end{itemize}

The confusion matrix is a table with two rows and two columns that reports the number of false positives, false negatives, true positives, and true negatives.

\begin{figure}[h!]
    \begin{center}
        \begin{tikzpicture}[
                box/.style={draw,rectangle,minimum size=2cm,text width=1.5cm,align=left}]
            \matrix (conmat) [row sep=.1cm,column sep=.1cm] {
                \node (tpos) [box,
                    label=left:\( \mathbf{P'} \),
                    label=above:\( \mathbf{P} \),
                ] {True \\ positive \textbf{TP}};
                 &
                \node (fneg) [box,
                    label=above:\textbf{n},
                    label=above right:\textbf{total},
                    label=right:\( \mathrm{P}' \)
                ]{False \\ negative \textbf{FN}};
                \\
                \node (fpos) [box,
                    label=left:\( \mathbf{N'} \),
                    label=below left:\textbf{total},
                    label=below:P
                ] {False \\ positive \textbf{FP}};
                 &
                \node (tneg) [box,
                    label=right:\( \mathrm{N}' \),
                    label=below:N
                ] {True \\ negative \textbf{TN}};
                \\
            };
            \node [left=.05cm of conmat,text width=1.5cm,align=right] {\textbf{actual \\ value}};
            \node [above=.05cm of conmat] {\textbf{prediction outcome}};
        \end{tikzpicture}
        \caption{Confusion matrix description}
    \end{center}
\end{figure}

The matrix summarizes the predictions from a classification model, i.e.,
how well the model performed when predicting the class labels for positive
and negative samples.
While the matrix presents the most informative view of the performance of the model,
we still need to summarize the information in the matrix into a single number(s)
that can be used to compare different models.

In this paper, we will use the following metrics to compare the performance of the anomaly detection models:

\begin{itemize}
    \item \textbf{Accuracy} is the fraction of predictions that the model got right.
          It is defined as follows:
          $$\text{Accuracy}=\frac{\textbf{TP}+\textbf{TN}}{\textbf{TP}+\textbf{TN}+\textbf{FP}+\textbf{FN}}$$
          While this metric is easy to understand, it is not very informative when the dataset is imbalanced
          which is the case for the anomaly detection task where the proportion
          of positive samples is low.

    \item \textbf{Precision} is the fraction of positive predictions that were correct.
          $$\text{Precision}=\frac{\textbf{TP}}{\textbf{TP}+\textbf{FP}}$$
          This metric is useful when the cost of false positives is high.
          For example, in the case of anomaly detection, we want to have
          a high precision so that we do not have to manually check many false positives
          or trigger any downstream filtering task too often.

    \item \textbf{Recall} is the fraction of positive samples that were correctly predicted.
          $$\text{Recall}=\frac{\textbf{TP}}{\textbf{TP}+\textbf{FN}}$$
          This metric is useful when the cost of false negatives is high.
          For example, in the case of anomaly detection, we want to minimize the number of false negatives
          because we do not want to miss any anomalies.

    \item \textbf{F1} is the harmonic mean of precision and recall which ranges from 0 to 1.
          $$\text{F1}=2\cdot\frac{\text{Precision}\cdot\text{Recall}}{\text{Precision}+\text{Recall}}$$
          This metric is useful when we want to balance the precision and recall.
          For example, in the case of anomaly detection, we want to have a high precision
          so that we do not have to manually check many false positives
          or trigger any downstream filtering task too often.
          At the same time, we want to minimize the number of false negatives
          because we do not want to miss any anomalies.

          The advantage of using this metric instead of the accuracy is that it
          can be used even when the dataset is highly imbalanced and it would
          detect if the model performs poorly in terms of precision and/or recall.
\end{itemize}

\subsubsection{Comparison}


\begin{table}[h!]
    \centering
    \begin{tabular}{|c|c|c|c|c|c|}
        \hline
        Dataset & Model              & Accuracy  & Precision & Recall    & F1        \\
        \hline
        KPI     & Linear Regression  & 0.61/0.66 & 0.07/0.15 & 0.73/0.69 & 0.12/0.25 \\
                & Transformer        & 0.98/0.97 & 0.89/0.95 & 0.59/0.63 & 0.71/0.76 \\
                & Linear Transformer & 0.97/0.95 & 0.58/0.73 & 0.62/0.69 & 0.60/0.71 \\
        \hline
        NAB     & Linear Regression  & 0.09/0.20 & 0.09/0.20 & 1.00/1.00 & 0.16/0.33 \\
                & Transformer        & 0.76/0.75 & 0.16/0.41 & 0.39/0.55 & 0.23/0.47 \\
                & Linear Transformer & 0.53/0.56 & 0.13/0.27 & 0.76/0.71 & 0.22/0.40 \\
        \hline
        FI2010  & Linear Regression  & 0.14/0.16 & 0.14/0.16 & 1.00/1.00 & 0.24/0.27 \\
                & Transformer        & 0.84/0.80 & 0.17/0.21 & 0.03/0.11 & 0.06/0.14 \\
                & Linear Transformer & 0.82/0.77 & 0.15/0.14 & 0.06/0.09 & 0.09/0.11 \\
        \hline
    \end{tabular}
    \caption{Final metrics. The reported value are for the train and validation set.}
    \label{tab:final_metrics}
\end{table}


The final metrics for the models are presented in Table \ref{tab:final_metrics}.
Refer to the Appendix \ref{app:training_plots} for the training plots for the models.

The Transformer model outperforms the other models on all datasets. And that
is expected because the Transformer model is more complex (i.e., has more parameters)
than the other models. However, the Linear Transformer model performs almost as well
and it provides better performance-resource utilization tradeoff (see Section~\ref{sec:performance}).

In terms of training robustness (i.e., how stable the training is, refer to Section~\ref{sec:training}),
a simple Linear Regression model is the most stable model, as expected. Both Transformer models suffer
from the gradient explosion problem which leads to unstable training
which can be partially remediated (see Section~\ref{sec:gradient_clipping}).

However, the performance of a simple Linear regression depends on the dataset. For example,
the Linear regression is not able to learn the patterns in the NAB dataset
and is stuck at predicting only the positive samples (i.e., anomalies).
As for the KPI dataset, the performance on the KPI can be tweaked by training
the model for a different number of epochs to achieve the required balance between
precision and recall (refer to Figure \ref{fig:LinearRegression_KPI_training}).
Both Transformer models, on the other hand, perform equally well on both datasets
as long as there is no gradient explosion problem.

\textbf{In short}, Transformer models outperform the simple linear regression
but can have unstable training. Linear Transformer model is slightly behind
the vanilla model. This which shows that untuned Transformer models
provide a better baseline for anomaly detection than the simple linear regression model.


\subsection{Performance/Speed} \label{sec:performance}

The performance of FPGA implementation of the models will be compared with the
pytorch CPU and GPU implementation of the models.
The CPU used for the experiments is Intel(R) Core(TM) i7-8750H CPU @ 2.20GHz clocked
to 4.20GHz and the GPU is NVIDIA GeForce GTX 1050 Ti.
The latency on CPU and GPU was measured empirically by running the inference
multiple times (10,000 runs).
Refer to the Appendix \ref{app:empirical_latency} for the histograms of
inference times.

\begin{table}[h]
    \centering
    \begin{tabular}{|c|c|c|}
        \hline
        Model             & Device & Inference time (microseconds) \\
        \hline
        Transformer       & CPU    & 353.35(113.50)                \\
                          & GPU    & 626.34(323.18)                \\
                          & FPGA   & 37.14                         \\
        \hline
        LinearTransformer & CPU    & 365(114.21)                   \\
                          & GPU    & 472.44(110.83)                \\
                          & FPGA   & 29.86                         \\
        \hline
    \end{tabular}
    \caption{Inference time for the models on different devices.
        For CPU and GPU we report the mean and standard deviation.
        For FPGA, the deterministic latency is reported (see Section~\ref{sec:transformer_optimizations}).
    }
    \label{tab:inference_time}
\end{table}

According to the results (Table \ref{tab:inference_time}), the FPGA implementation of the models outperforms
the CPU implementation by a large margin (achieving a speedup of 10x).
The reason for this is that the FPGA implementation is highly parallelized
compared to the CPU implementation.
Moreover, the benefit of using the FPGA implementation is that it has deterministic
latency (i.e., the inference time is constant) which is not the case for the CPU and GPU implementation
where the inference time varies depending on the load of the CPU/GPU (see the
standard deviation for the inference times of CPU and GPU).

\begin{remark}
    The GPU implementation of the models is slower than the CPU implementation.
    This is because the models are small and the overhead of transferring the data
    to the GPU is larger than the time it takes to perform the inference on the CPU
    for a single sample.
\end{remark}

Regarding the resource utilization, the Linear Transformer model uses more resources
than the Transformer model (see Table \ref{tab:resource_utilization}) and achieves
a slightly better runtime performance (see Table \ref{tab:inference_time}).

\begin{table}[h]
    \centering
    \begin{tabular}{|c|c|c|c|}
        \hline
        Model             & BRAM & FF   & LUT  \\
        \hline
        Transformer       & 5\%  & 30\% & 90\% \\
        \hline
        LinearTransformer & 18\% & 43\% & 99\% \\
        \hline
    \end{tabular}
    \caption{Relative resource utilization for the PYNQ-Z2 board.}
    \label{tab:resource_utilization}
\end{table}

\section{Conclusion} \label{sec:conclusion}

\subsection{Summary}

In this work, we have examined the performance of the Transformer model
for the anomaly detection task and how it can be effectively implemented on
FPGA.

We have shown how different Transformer Architectures (and especially the attention module)
affect the performance of the model both in terms of accuracy and inference speed.
It was shown that the Transformer model outperforms the simple linear regression model
without too much hyperparameter tuning. The only downside of the Transformer model
that was observed was the unstable training which can be partially remediated
by using gradient clipping.

We have also demonstrated how different optimizations affect the performance
and resource utilization of the model on FPGA.
We also provided an optimized implementation of the Transformer model on FPGA
which can be used as a reference.

This work can be used as a starting point for further research on the topic of
anomaly detection using Transformer models and for further research on the topic
of efficient implementation of Transformer models on FPGA in different domains
(e.g., natural language processing).

\subsection{Future work}
For future work, we propose the following:
\begin{enumerate}
    \item To try bigger FPGA boards to push the limit of maximum performance.
          The current implementations hit the limit of the PYNQ-Z2 board.
    \item To evaluate the performance on more recent \textbf{labeled} financial market data.
          Currently, there is no publicly available labeled financial market dataset
          with anomalies. The FI2010 dataset is not labeled and we had to inject synthetic anomalies.
    \item To evaluate the performance Transformer on FPGA board for other tasks (e.g., natural language processing).
          The current implementation is only for the anomaly detection task.
          The other domains might involve a slightly different architecture
          (e.g., different number of layers, different number of heads, etc.),
          it can be valuable to have those models on edge devices because of their
          recent surge in popularity and increased use.
          Currently, the language models are too big to fit on the FPGA board (especially, PYNQ-Z2),
          however, the smaller models can still be implemented and evaluated on bigger boards.
\end{enumerate}

\bibliographystyle{unsrt}

\begin{thebibliography}{10}

\bibitem{Falkenberry_2008}
Thomas Neil~Falkenberry CFA.
\newblock High frequency data filtering.
\newblock https://www.tickdata.com/whitepaper/high-frequency-data-filtering, Sep 2008.

\bibitem{Vallis_Hochenbaum_Twitter}
Owen Vallis, Jordan Hochenbaum, and Twitter.
\newblock Introducing practical and robust anomaly detection in a time series.
\newblock \url{https://blog.twitter.com/engineering/en_us/a/2015/introducing-practical-and-robust-anomaly-detection-in-a-time-series}.

\bibitem{AnomalyDataBig}
Mohiuddin Ahmed, Nazim Choudhury, and Shahadat Uddin.
\newblock Anomaly detection on big data in financial markets.
\newblock In {\em 2017 IEEE/ACM International Conference on Advances in Social Networks Analysis and Mining (ASONAM)}, pages 998--1001, 2017.

\bibitem{TransformersNLP}
Anthony Gillioz, Jacky Casas, Elena Mugellini, and Omar~Abou Khaled.
\newblock Overview of the transformer-based models for nlp tasks.
\newblock In {\em 2020 15th Conference on Computer Science and Information Systems (FedCSIS)}, pages 179--183, 2020.

\bibitem{2202.07125}
Qingsong Wen, Tian Zhou, Chaoli Zhang, Weiqi Chen, Ziqing Ma, Junchi Yan, and Liang Sun.
\newblock Transformers in time series: A survey.
\newblock 2022.

\bibitem{10.1007/978-3-319-56258-2_14}
Andreea-Ingrid Funie, Liucheng Guo, Xinyu Niu, Wayne Luk, and Mark Salmon.
\newblock Custom framework for run-time trading strategies.
\newblock In Stephan Wong, Antonio~Carlos Beck, Koen Bertels, and Luigi Carro, editors, {\em Applied Reconfigurable Computing}, pages 154--167, Cham, 2017. Springer International Publishing.

\bibitem{1905.05667}
Julio-Omar Palacio-Niño and Fernando Berzal.
\newblock Evaluation metrics for unsupervised learning algorithms, 2019.

\bibitem{Steinbuss2020GeneratingAO}
Georg Steinbuss and Klemens B{\"o}hm.
\newblock Generating artificial outliers in the absence of genuine ones — a survey.
\newblock {\em ACM Transactions on Knowledge Discovery from Data (TKDD)}, 15:1 -- 37, 2020.

\bibitem{Schmidl2022AnomalyDI}
Sebastian Schmidl, Phillip Wenig, and Thorsten Papenbrock.
\newblock Anomaly detection in time series: A comprehensive evaluation.
\newblock {\em Proc. VLDB Endow.}, 15:1779--1797, 2022.

\bibitem{1409.0473}
Dzmitry Bahdanau, Kyunghyun Cho, and Yoshua Bengio.
\newblock Neural machine translation by jointly learning to align and translate, 2014.

\bibitem{1706.03762}
Ashish Vaswani, Noam Shazeer, Niki Parmar, Jakob Uszkoreit, Llion Jones, Aidan~N. Gomez, Lukasz Kaiser, and Illia Polosukhin.
\newblock Attention is all you need, 2017.

\bibitem{Devlin2019BERTPO}
Jacob Devlin, Ming-Wei Chang, Kenton Lee, and Kristina Toutanova.
\newblock Bert: Pre-training of deep bidirectional transformers for language understanding.
\newblock {\em ArXiv}, abs/1810.04805, 2019.

\bibitem{Jeong2023AnomalyBERTST}
Yungi Jeong, Eu-Hui Yang, Jung~Hyun Ryu, Imseong Park, and Myung joo Kang.
\newblock Anomalybert: Self-supervised transformer for time series anomaly detection using data degradation scheme.
\newblock {\em ArXiv}, abs/2305.04468, 2023.

\bibitem{Dufter2021PositionII}
Philipp Dufter, Martin Schmitt, and Hinrich Sch{\"u}tze.
\newblock Position information in transformers: An overview.
\newblock {\em Computational Linguistics}, 48:733--763, 2021.

\bibitem{Weng_2023}
Lilian Weng.
\newblock The transformer family version 2.0.
\newblock \url{https://lilianweng.github.io/posts/2023-01-27-the-transformer-family-v2}, Jan 2023.

\bibitem{1512.03385}
Kaiming He, Xiangyu Zhang, Shaoqing Ren, and Jian Sun.
\newblock Deep residual learning for image recognition, 2015.

\bibitem{1607.06450}
Jimmy~Lei Ba, Jamie~Ryan Kiros, and Geoffrey~E. Hinton.
\newblock Layer normalization, 2016.

\bibitem{2006.16236}
Angelos Katharopoulos, Apoorv Vyas, Nikolaos Pappas, and François Fleuret.
\newblock Transformers are rnns: Fast autoregressive transformers with linear attention, 2020.

\bibitem{tuli2022tranad}
Shreshth Tuli, Giuliano Casale, and Nicholas~R Jennings.
\newblock {TranAD: Deep Transformer Networks for Anomaly Detection in Multivariate Time Series Data}.
\newblock {\em Proceedings of VLDB}, 15(6):1201--1214, 2022.

\bibitem{2205.13504}
Ailing Zeng, Muxi Chen, Lei Zhang, and Qiang Xu.
\newblock Are transformers effective for time series forecasting?, 2022.

\bibitem{Lezmi2023TimeSF}
Edmond Lezmi and Jiali Xu.
\newblock Time series forecasting with transformer models and application to asset management.
\newblock {\em SSRN Electronic Journal}, 2023.

\bibitem{VitisYork}
University of~York.
\newblock {Vitis HLS Knowledge Base - Real-Time Systems - York Wiki Service}.
\newblock \url{https://wiki.york.ac.uk/display/RTS/Vitis+HLS+Knowledge+Base}, July 2020.

\bibitem{AMD2023VitisHLS}
Xilinx Inc.
\newblock Vitis {High}-{Level} {Synthesis} {User} {Guide}.
\newblock \url{https://docs.xilinx.com/r/en-US/ug1399-vitis-hls/Design-Principles}, May 2023.

\bibitem{AMD2023VitisSimCosim}
Xilinx Inc.
\newblock {C/RTL Co-Simulation in Vitis HLS}.
\newblock \url{https://docs.xilinx.com/r/en-US/ug1399-vitis-hls/C/RTL-Co-Simulation-in-Vitis-HLS}, July 2023.

\bibitem{AMD2023Pipelining}
Xilinx Inc.
\newblock Vitis hls: Pipelining loops.
\newblock \url{https://docs.xilinx.com/r/en-US/ug1399-vitis-hls/Design-Principles}, May 2023.

\bibitem{intelpipelineloopsII}
Intel.
\newblock Intel high level synthesis: Best practices guide.
\newblock \url{https://www.intel.com/content/www/us/en/docs/programmable/683152/21-3/pipeline-loops.html}.

\bibitem{1506.01186}
Leslie~N. Smith.
\newblock Cyclical learning rates for training neural networks, 2015.

\bibitem{Kingma2014AdamAM}
Diederik~P. Kingma and Jimmy Ba.
\newblock Adam: A method for stochastic optimization.
\newblock {\em CoRR}, abs/1412.6980, 2014.

\bibitem{lightning.ai_2023_lr_finder}
lightning.ai.
\newblock Learningratefinder — pytorch lightning 2.0.7 documentation.
\newblock \url{https://lightning.ai/docs/pytorch/stable/api/lightning.pytorch.callbacks.LearningRateFinder.html}, Aug 2023.

\bibitem{1211.5063}
Razvan Pascanu, Tomas Mikolov, and Yoshua Bengio.
\newblock On the difficulty of training recurrent neural networks, 2012.

\bibitem{Good_1952}
I.~J. Good.
\newblock Rational decisions.
\newblock {\em Journal of the Royal Statistical Society: Series B (Methodological)}, 14(1):107–114, Jan 1952.

\bibitem{scikit-learn}
F.~Pedregosa, G.~Varoquaux, A.~Gramfort, V.~Michel, B.~Thirion, O.~Grisel, M.~Blondel, P.~Prettenhofer, R.~Weiss, V.~Dubourg, J.~Vanderplas, A.~Passos, D.~Cournapeau, M.~Brucher, M.~Perrot, and E.~Duchesnay.
\newblock Scikit-learn: Machine learning in {P}ython.
\newblock {\em Journal of Machine Learning Research}, 12:2825--2830, 2011.

\bibitem{Ahmad2017Unsupervised}
Subutai Ahmad, Alexander Lavin, Scott Purdy, and Zuha Agha.
\newblock Unsupervised real-time anomaly detection for streaming data.
\newblock {\em Neurocomputing}, 262:134--147, 11 2017.
\newblock [Online; accessed 2023-07-19].

\bibitem{2208.03938}
Zeyan Li, Nengwen Zhao, Shenglin Zhang, Yongqian Sun, Pengfei Chen, Xidao Wen, Minghua Ma, and Dan Pei.
\newblock Constructing large-scale real-world benchmark datasets for aiops, 2022.

\bibitem{1705.03233}
Adamantios Ntakaris, Martin Magris, Juho Kanniainen, Moncef Gabbouj, and Alexandros Iosifidis.
\newblock Benchmark dataset for mid-price forecasting of limit order book data with machine learning methods.
\newblock 2017.

\bibitem{Crepey2022Anomaly}
St{\' e}phane Cr{\' e}pey, Noureddine Lehdili, Nisrine Madhar, and Maud Thomas.
\newblock Anomaly {Detection} in {Financial} {Time} {Series} by {Principal} {Component} {Analysis} and {Neural} {Networks}.
\newblock {\em Algorithms}, 15(10):385, oct 19 2022.
\newblock [Online; accessed 2023-07-19].

\end{thebibliography}

\newpage
\appendix
\section{Training plots} \label{app:training_plots}

\begin{figure}[h!]
    \centering
    \caption{Training metrics for different epochs for Transformer model on KPI dataset}
    \includegraphics[width=\textwidth]{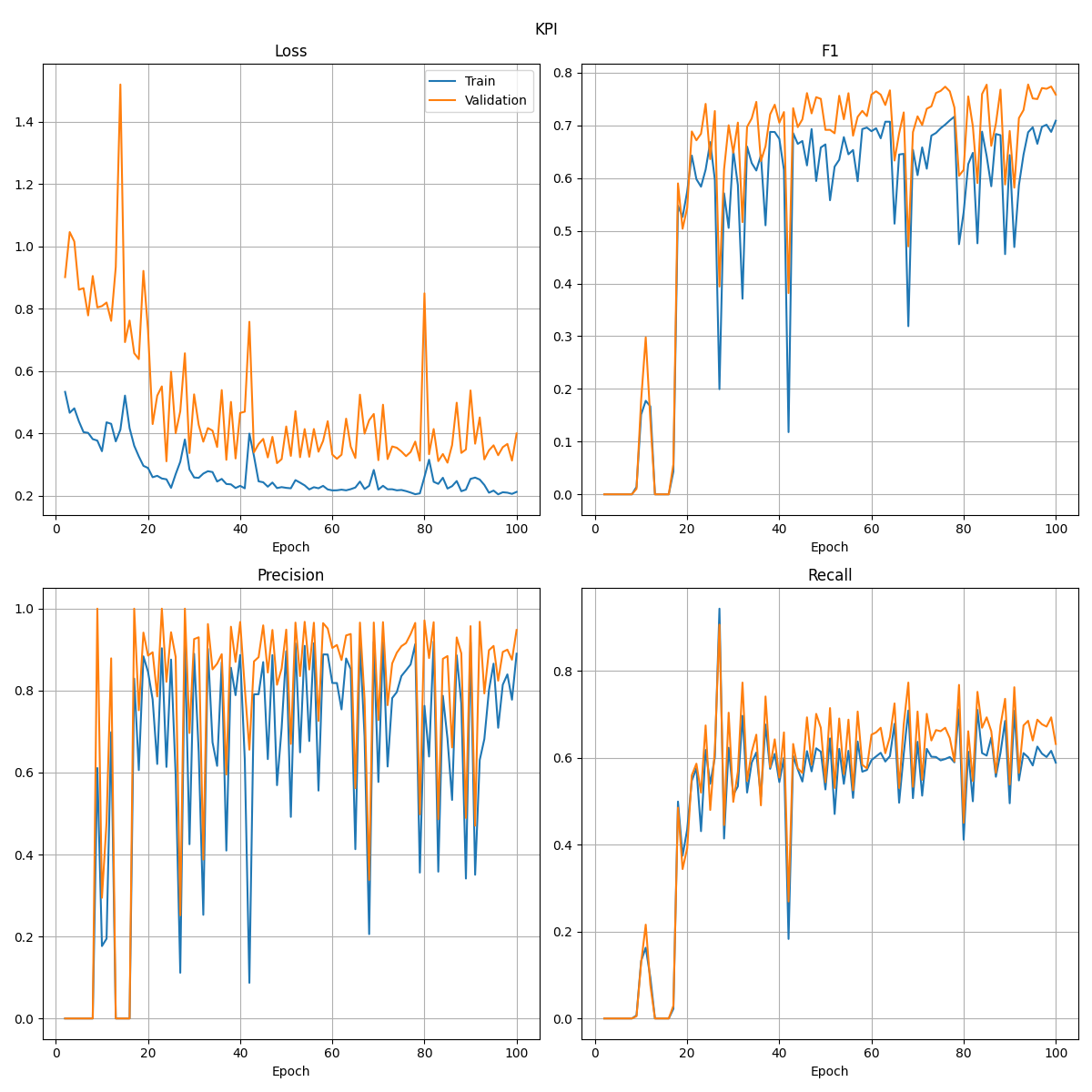}
    \label{fig:Transformer_KPI_training}
\end{figure}

\begin{figure}[h!]
    \centering
    \caption{Training metrics for different epochs for Transformer model on NAB dataset}
    \includegraphics[width=\textwidth]{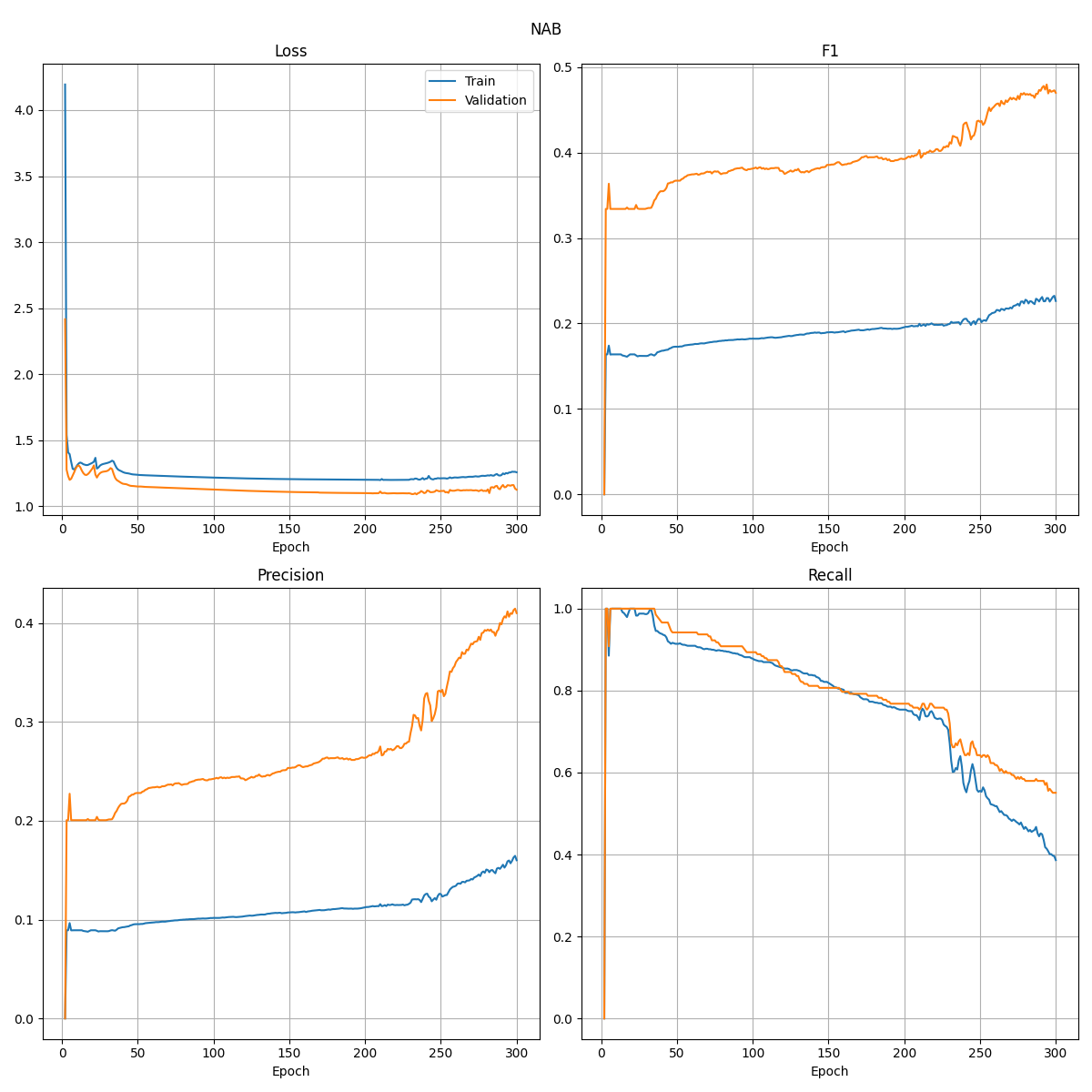}
    \label{fig:Transformer_NAB_training}
\end{figure}

\begin{figure}[h!]
    \centering
    \caption{Training metrics for different epochs for Linear Transformer model on KPI dataset}
    \includegraphics[width=\textwidth]{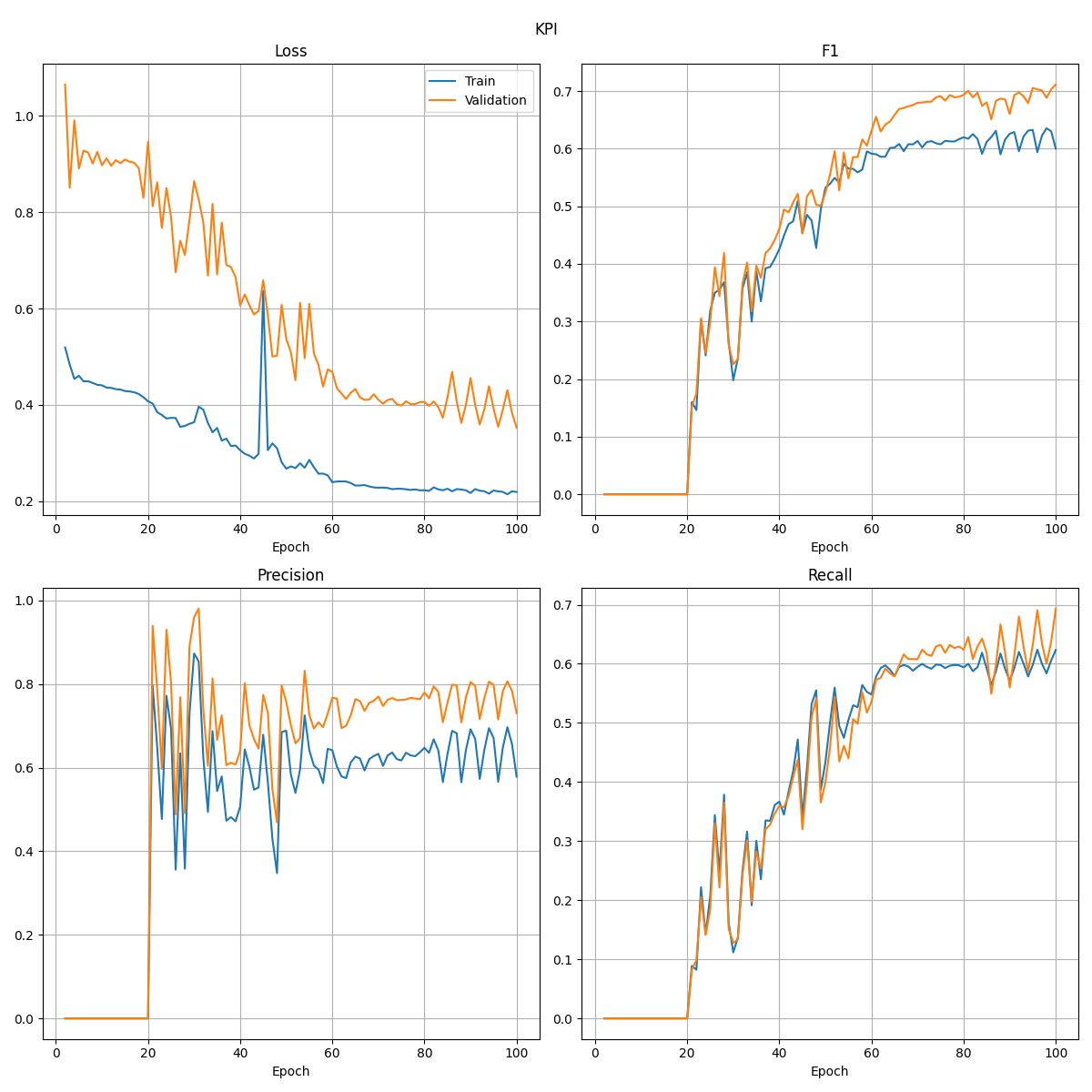}
    \label{fig:LinearTransformer_KPI_training}
\end{figure}

\begin{figure}[h!]
    \centering
    \caption{Training metrics for different epochs for Linear Transformer model on NAB dataset}
    \includegraphics[width=\textwidth]{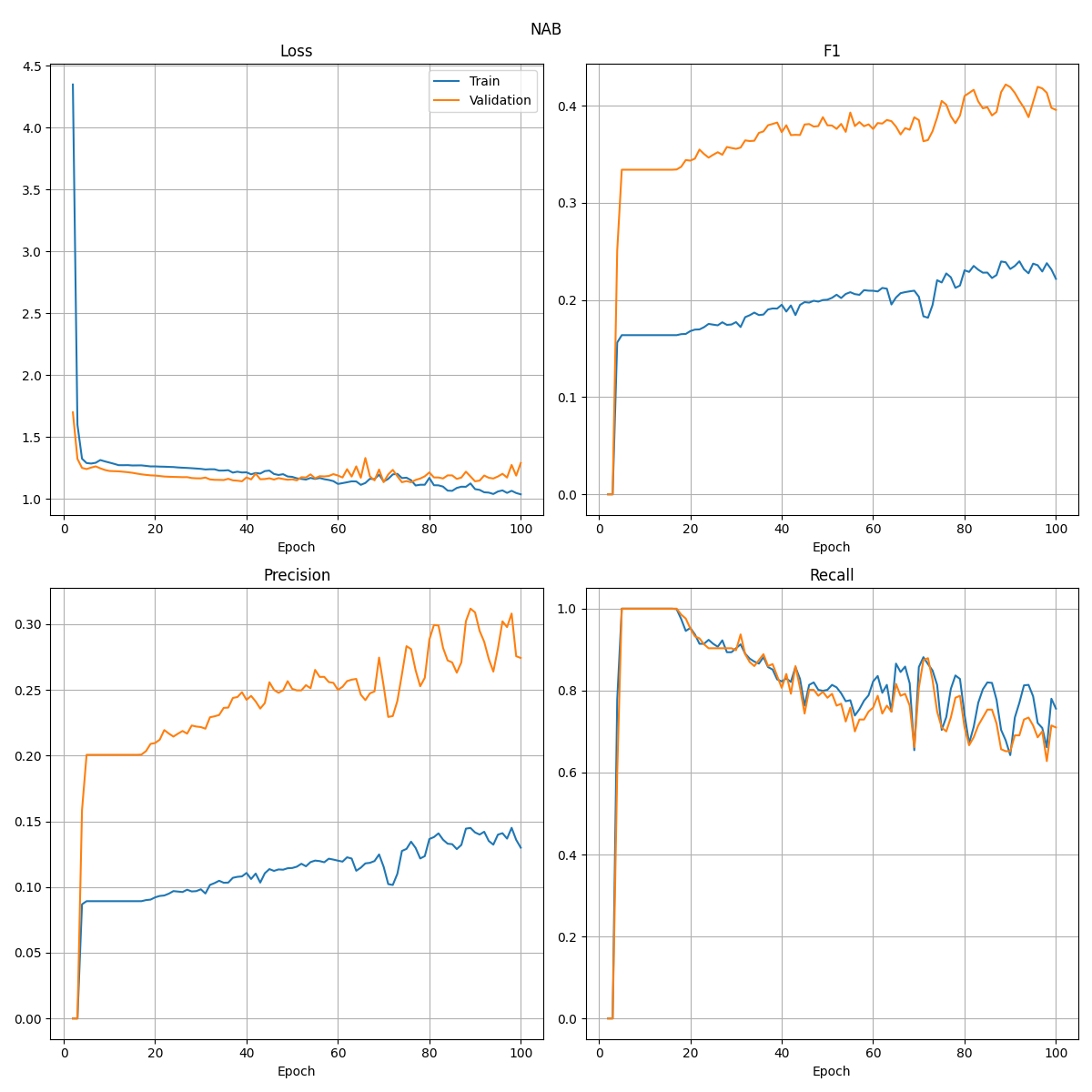}
    \label{fig:LinearTransformer_NAB_training}
\end{figure}

\begin{figure}[h!]
    \centering
    \caption{Training metrics for different epochs for Linear Regression model on KPI dataset}
    \includegraphics[width=\textwidth]{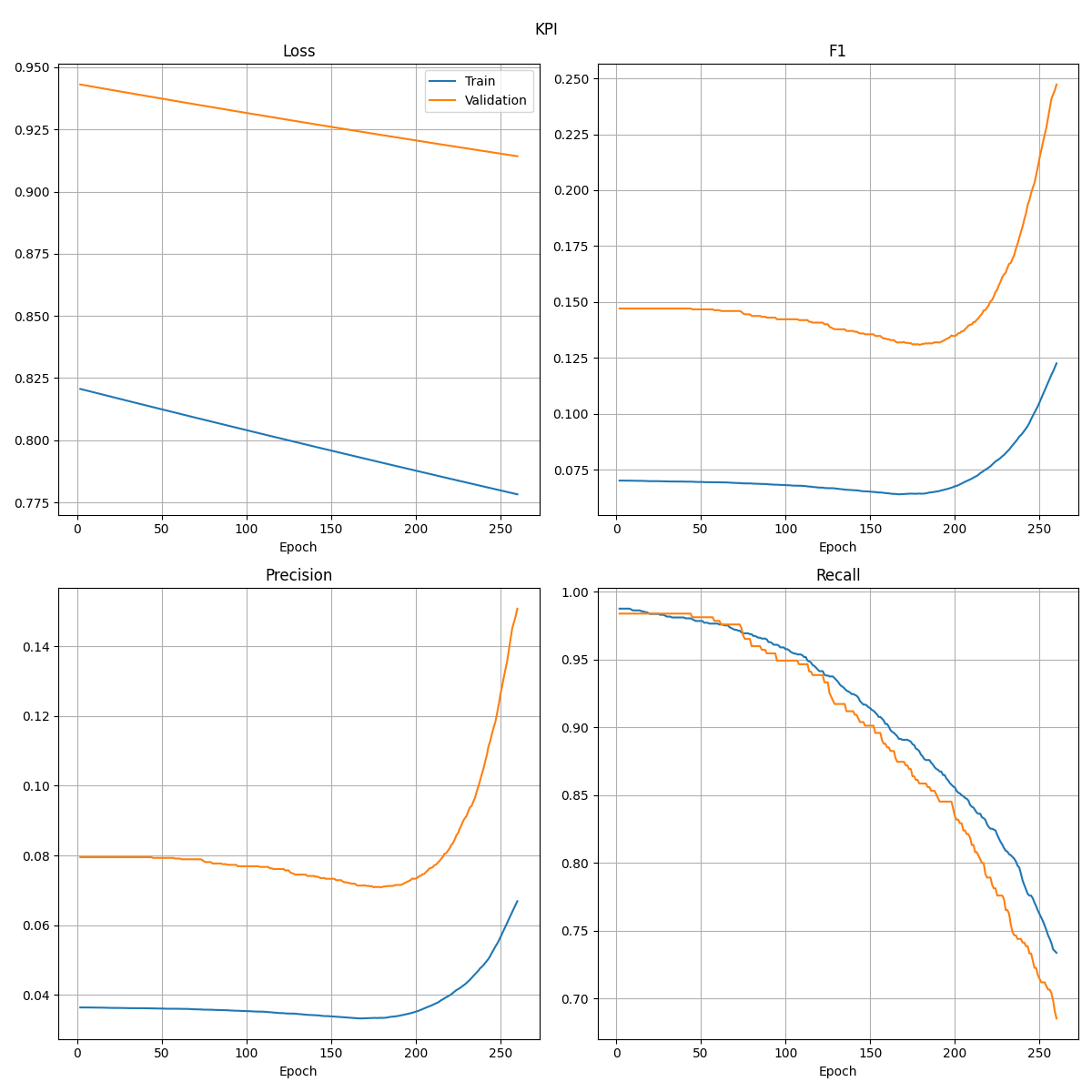}
    \label{fig:LinearRegression_KPI_training}
\end{figure}

\begin{figure}[h!]
    \centering
    \caption{Training metrics for different epochs for Linear Regression model on NAB dataset}
    \includegraphics[width=\textwidth]{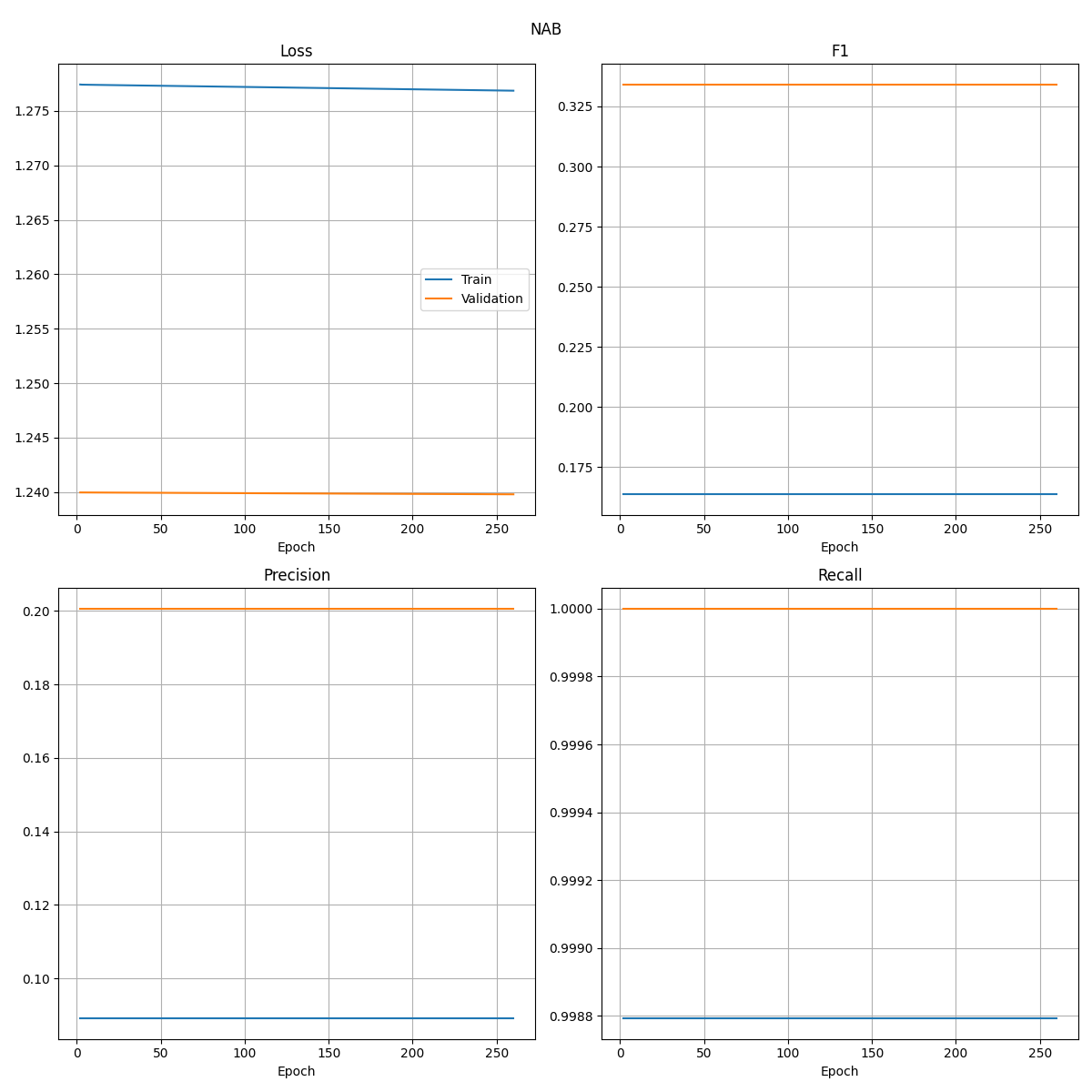}
    \label{fig:LinearRegression_NAB_training}
\end{figure}

The training plots are presented in Figures~\ref{fig:Transformer_KPI_training}, \ref{fig:Transformer_NAB_training}, \ref{fig:LinearTransformer_KPI_training}, \ref{fig:LinearTransformer_NAB_training}, \ref{fig:LinearRegression_KPI_training}, \ref{fig:LinearRegression_NAB_training}.

\section{Empirical latency} \label{app:empirical_latency}

\begin{figure}[h!]
    \centering
    \caption{Empirical latency for Transformer model on CPU.}
    \includegraphics[width=\textwidth]{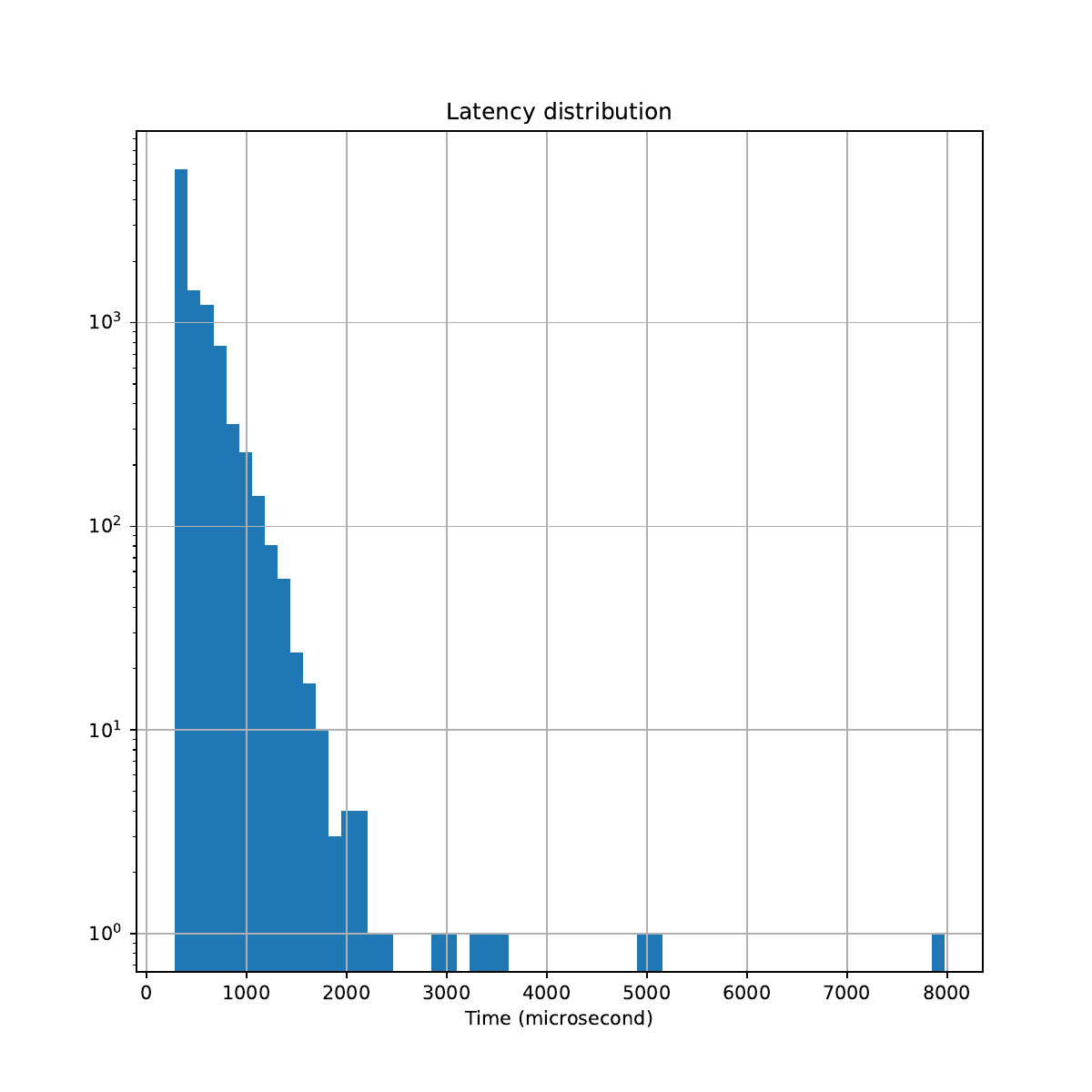}
    \label{fig:transformer_cpu_latency}
\end{figure}

\begin{figure}[h!]
    \centering
    \caption{Empirical latency for Transformer model on GPU.}
    \includegraphics[width=\textwidth]{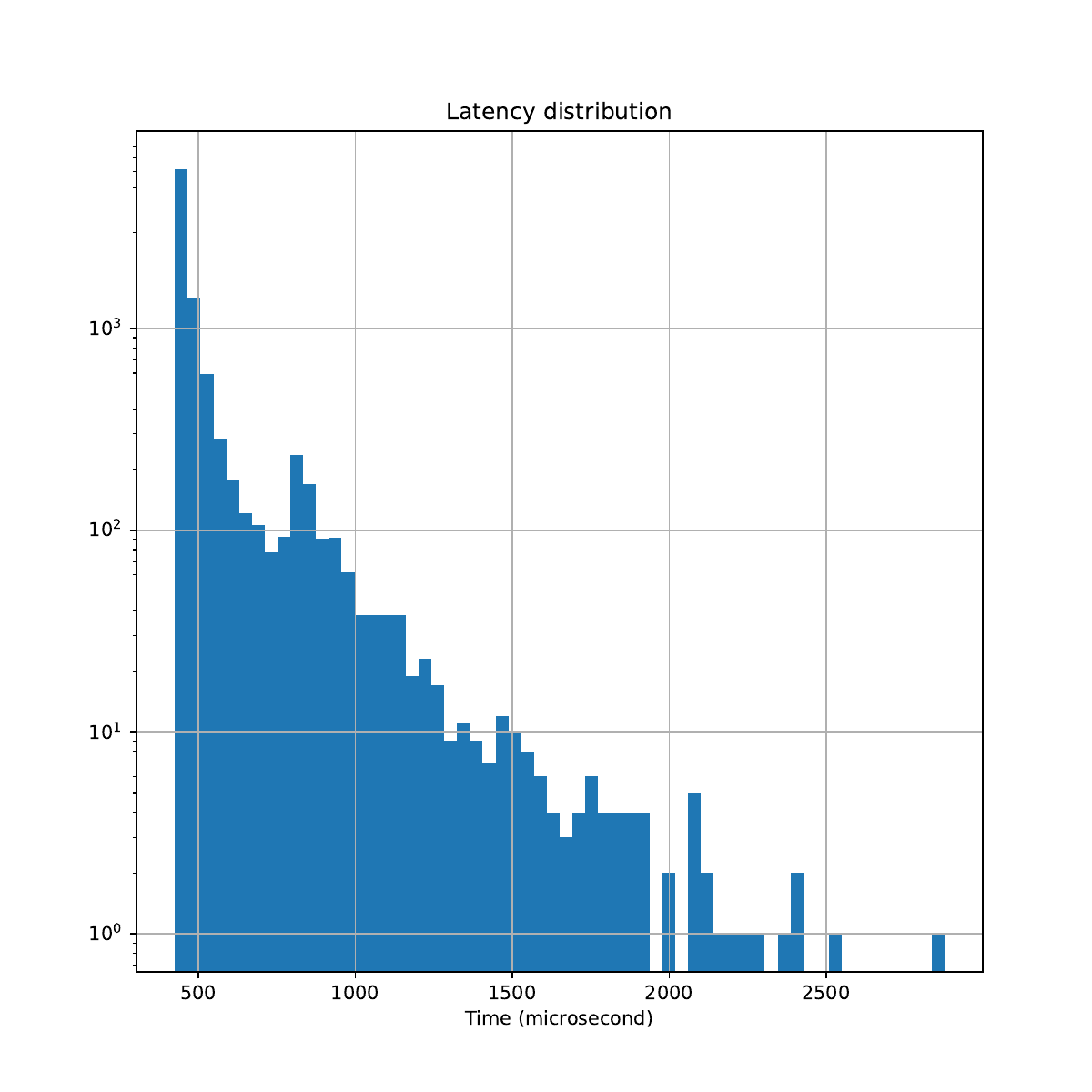}
    \label{fig:transformer_gpu_latency}
\end{figure}

\begin{figure}[h!]
    \centering
    \caption{Empirical latency for Linear Transformer model on CPU.}
    \includegraphics[width=\textwidth]{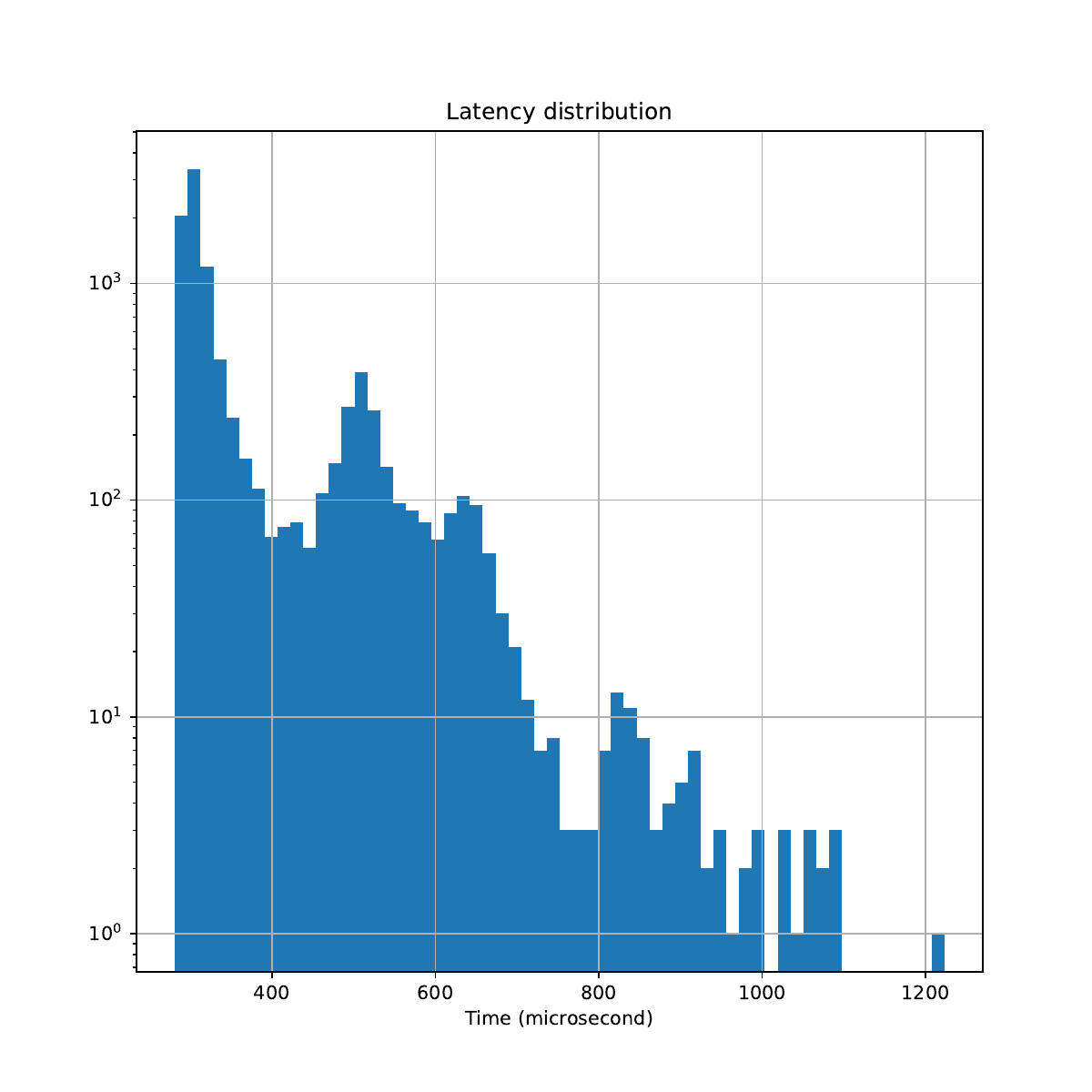}
    \label{fig:linear_transformer_cpu_latency}
\end{figure}

\begin{figure}[h!]
    \centering
    \caption{Empirical latency for Linear Transformer model on GPU.}
    \includegraphics[width=\textwidth]{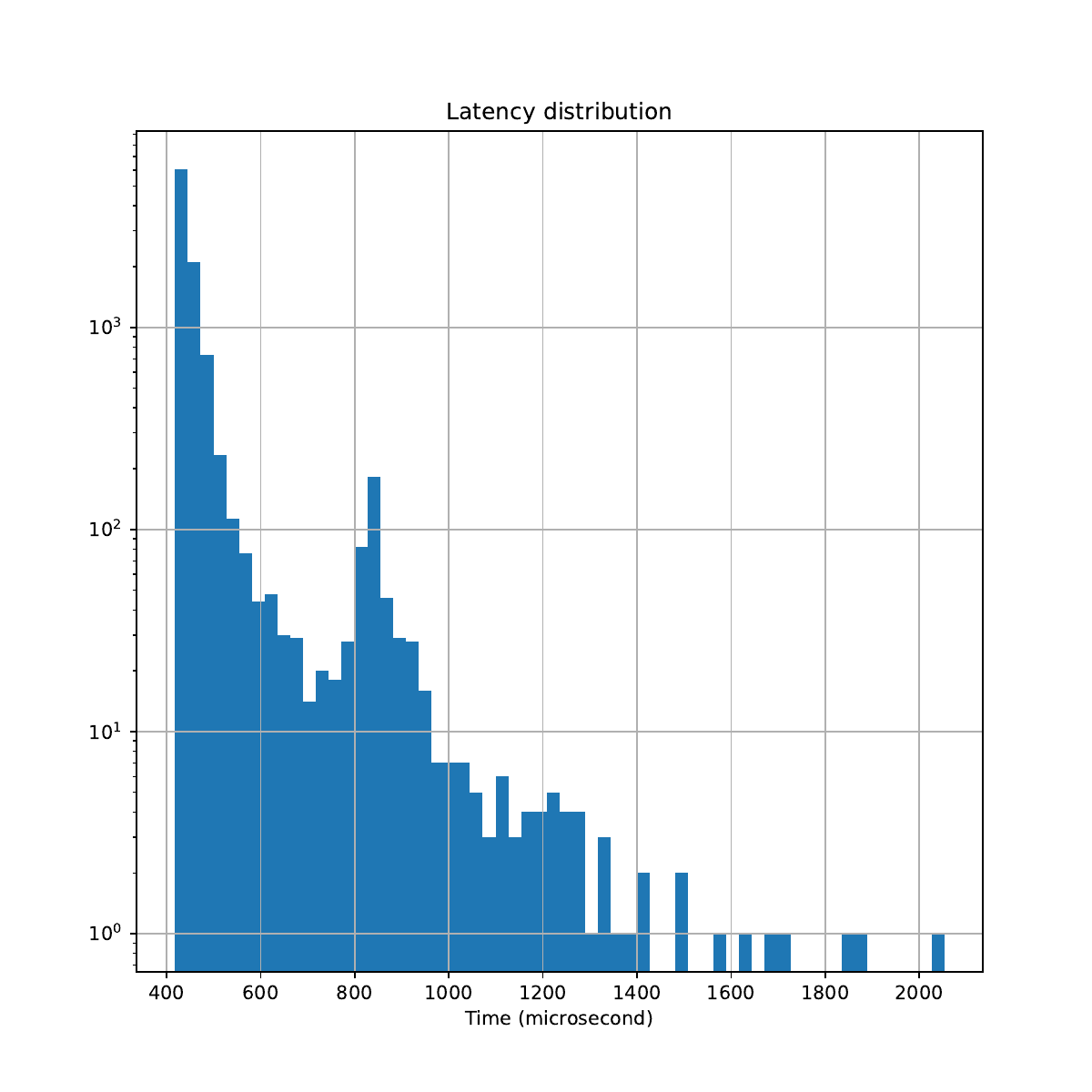}
    \label{fig:linear_transformer_gpu_latency}
\end{figure}

The empirical latency plots are presented in Figures~\ref{fig:transformer_cpu_latency}, \ref{fig:transformer_gpu_latency}, \ref{fig:linear_transformer_cpu_latency}, \ref{fig:linear_transformer_gpu_latency}.

\end{document}